\pdfoutput=1

\documentclass[11pt]{article}

\usepackage[final]{acl}

\usepackage{times}
\usepackage{latexsym}
\usepackage{float} 
\usepackage[T1]{fontenc}

\usepackage[utf8x]{inputenc}
\usepackage{hyperref}
\usepackage{microtype}

\usepackage{inconsolata}

\usepackage{graphicx}

\usepackage{pgfplots}

\usepackage{multirow}
\usepackage{xcolor,colortbl}
\usepackage{makecell}

\begin{document}

\title{From MTEB to MTOB: \\ Retrieval-Augmented Classification for Descriptive Grammars}


\author{Albert Kornilov \\
  HSE University \\
  Sberbank \\ \\\And
  Tatiana Shavrina \\
  Institute of Linguistics RAS \\
 \\}

\maketitle
\begin{abstract}
Recent advances in language modeling have demonstrated significant improvements in zero-shot capabilities, including in-context learning, instruction following, and machine translation for extremely under-resourced languages \citep{tanzer2024benchmarklearningtranslatenew}. However, many languages with limited written resources rely primarily on descriptions of grammar and vocabulary.

In this paper, we introduce a set of benchmarks to evaluate how well models can extract and classify information from the complex descriptions found in linguistic grammars. We present a Retrieval-Augmented Generation (RAG)-based approach that leverages these descriptions for downstream tasks such as machine translation. Our benchmarks encompass linguistic descriptions for 248 languages across 142 language families, focusing on typological features from WALS \citep{wals} and Grambank \citep{grambank_release}.

This set of benchmarks offers the first comprehensive evaluation of language models’ in-context ability to accurately interpret and extract linguistic features, providing a critical resource for scaling NLP to low-resource languages. The code and data are publicly available at \url{https://github.com/al-the-eigenvalue/RAG-on-grammars}.

\end{abstract}

\section{Introduction}
The advent of text-based foundational models has accelerated advancements in natural language processing, enhancing multilingual capabilities and applied tasks such as zero-shot machine translation, reading comprehension, and information extraction. Innovations like Machine Translation from One Book (MTOB) \citep{tanzer2024benchmarklearningtranslatenew} utilize descriptive grammars to improve translation performance for extremely low-resource languages, showing the potential of large-scale language models in bridging linguistic theory and practical NLP applications. Despite the promise, using descriptive grammars for zero-shot MT presents challenges like terminology variability, non-standard structures, and scattered relevant information.

Additionally, the Massive Text Embedding Benchmark (MTEB) \citep{muennighoff2023mtebmassivetextembedding} provides a thorough evaluation of text embeddings across diverse tasks and languages. A key challenge remains: the effective application of these models to descriptive grammars for languages with scant resources, typically supported by linguistic materials such as grammars and dictionaries.

This paper seeks to address these challenges by providing a systematic framework for extracting information from descriptive grammars and creating a scalable pipeline for descriptive grammar systematization. The key aspect of this approach is Retrieval Augmented Generation (RAG), which allows for the extraction of relevant information from grammars based on a specific typological characteristic (e.g., Order of Subject, Object and Verb). Based on the extracted paragraphs, an LLM determines the value of this characteristic (e.g., Subject-Verb-Object).

In this paper, we present the following contributions:

1. The first scaled linguistic evaluation of the LLM machine reading capabilities on descriptive grammars.

2. A pipeline based on Retrieval Augmented Generation (RAG), which extracts relevant paragraphs from grammars based on a given typological characteristic (for example, WALS 81A: Order of Subject, Object and Verb\footnote{\url{https://wals.info/feature/81A}}) and provides them as prompts to an LLM to determine the meaning of these characteristics (for example, Subject-Verb-Object). The pipeline is evaluated through extensive experiments.

3. A benchmark consisting of 700 paragraphs from 14 descriptive grammars, annotated according to whether a linguist can unambiguously determine information about word order (WALS 81A) in the language described, in order to evaluate the quality of information retrieval methods on the task of filtering out the irrelevant paragraphs separately from the RAG pipeline.

4. A benchmark for the RAG pipeline, consisting of 148 grammars for each feature, in order to assess LLMs’ capabilities of determining typological characteristics based on the entire grammar at once and evaluate the effectiveness of different combinations of RAG pipeline components (i.e., information retrieval methods and prompts).

The proposed framework, alongside the presented benchmarks, aims to contribute to the ongoing efforts to improve the quality and efficiency of machine translation systems and to aid linguists in typological research by semi-automating extraction of data from descriptive grammars and. The entire codebase is open-source, licensed under \texttt{MIT}, and available at \url{https://anonymous.4open.science/r/from-MTEB-to-MTOB}.

\section{Related Work}

Several advances in language modeling outline the possibility to leverage linguistic descriptions: retrieval-augmented generation methods for information extraction and generation, in-context abilities to operate with the descriptive texts in the prompt, and the availability of the texts in the machine-readable format.
\subsection{Retrieval-Augmented Generation}
Retrieval-Augmented Generation (RAG) has become an effective method for augmenting large language models (LLMs) by integrating external retrieval mechanisms. Instead of relying solely on in-model knowledge, RAG enables models to retrieve relevant information from external documents during the generation process \citep{lewis2020retrieval}. This approach has shown promising results in a variety of tasks, such as open-domain question answering and document summarization, by improving the factual accuracy and extending the context of models \citep{izacard2021leveragingpassageretrievalgenerative}. In particular, RAG has been instrumental in improving the performance of LLMs for tasks involving sparse and domain-specific data, where it retrieves external knowledge to complement the model’s inherent capabilities.

RAG’s potential for handling low-resource languages and complex descriptions, such as linguistic grammars, has not been fully explored. Recent advances in retrieval-based frameworks point toward its applicability in linguistic resource-scarce domains, where available data is often fragmented and incomplete \citep{gao2024retrievalaugmentedgenerationlargelanguage}. This paper leverages the principles of RAG to enhance machine translation and language modeling for under-resourced languages through the use of descriptive grammars.

\subsection{Zero-shot Learning}
Zero-shot learning has become a key area of research in natural language processing (NLP), particularly with the advent of large-scale pre-trained models. The ability of models to generalize to new tasks and languages without explicit task-specific training data is critical for expanding NLP applications to low-resource languages \citep{brown2020languagemodelsfewshotlearners}. LLMs, such as GPT-3 and GPT-4, have demonstrated impressive zero-shot capabilities in tasks ranging from text classification to machine translation, making them an essential tool for under-resourced languages where labeled data is scarce \citep{raffel2023exploringlimitstransferlearning}.

However, current zero-shot models still struggle with languages that have extremely limited or no monolingual or bilingual corpora. Recent studies, such as \citep{tanzer2024benchmarklearningtranslatenew} and \citep{zhang2024hire}, have shown that integrating linguistic descriptions—such as those found in grammars—can significantly improve zero-shot performance for these languages. In this work, we aim to further this line of research by evaluating the ability of models to utilize descriptive grammars in zero-shot settings.

\subsection{Grammar Use in NLP}
The use of linguistic grammars in NLP has been relatively underexplored, with most efforts focusing on leveraging corpora and parallel data for model training. However, as \citep{Visser2022} and others have demonstrated, grammars provide a rich source of structured linguistic knowledge, especially for low-resource languages where corpora are unavailable. According to \citep{bapna2022building}, as cited in \citep{zhang2024hire}, 95\% of the world's known languages do not have enough data for LLM training (fewer than 100K sentences), while most languages have linguistic materials available: 60\% have a descriptive grammar, and 75\% have a dictionary. 

Prior work, such as \citep{erdmann2017low}, demonstrates that integrating linguistic knowledge into a translation model through morphosyntactic parsing can improve translation accuracy for low-resource languages and dialects. However, the challenge of incorporating grammars into NLP models lies in the complexity and formalism of linguistic materials. This paper strives to facilitate further use of grammars in NLP tasks, such as machine translation, through benchmarking the ability of large language models to understand grammars.

\subsection{Extracting Typological Features from Grammars}
Existing research regarding extraction of typological features from grammars precedes the advent of large language models, relying on rule-based methods and earlier developments in classical machine learning and deep learning. The series of works by Virk et al. \citep{virk2017automatic}; \citep{virk2019exploiting}; \citep{virk2020linguistic}, \citep{virk2021deep} utilize methods that require extensive annotation of semantic frames; \citep{hammarstrom2020term} present a method applicable to binary typological features only, and the framework proposed by \citep{kornilov-2023-multilingual} is limited to information retrieval. In this paper, we leverage state-of-the-art language models for the task of typological feature extraction and seek to demonstrate LLMs' capabilities on the domain of linguistic descriptions.

\section{Method and Overall Architecture}

The basic pipeline for retrieval augmented generation (Naive RAG) \citep{gao2024retrievalaugmentedgenerationlargelanguage} consists of a database of documents, a retriever, a process of combining the retrieved documents with the prompt, and the LLM generating an answer based on the prompt.

Advanced RAG pipelines described in \citep{gao2024retrievalaugmentedgenerationlargelanguage} are modifications of different parts of the Naive RAG pipeline. In the context of retrieval augmented generation from a descriptive grammar, the first component of the pipeline—the database of documents—is the grammar itself, hence it is fixed and not as modifiable as for RAG tasks that utilize the Internet or a large database for answering one question. We chose the chunking method that is simple but still context-aware—splitting grammars into paragraphs. We avoid model-based adaptive chunking methods, which would interfere with interpreting the subsequent RAG components due to grammars being a relatively underexplored domain.

The second component of the RAG pipeline is the retrieval method. We evaluate BM25 (described in \citep{trotman2012towards} and taken as a baseline in \citep{trotman2014improvements}), a language-agnostic retriever based on term frequency, and state-of-the-art retrievers/rerankers based on embeddings featured in the Massive Text Embedding Benchmark (MTEB) leaderboard \citep{muennighoff2023mtebmassivetextembedding}. Ideally, the chosen embedding-based retrieval methods should respond well to linguistic diversity, since descriptive grammars contain examples in the described language, which may contain diacritic signs and subwords or segments that are rare or unused in English. The weights assigned by the tokenizer to embeddings of such symbols due to their absence in the vocabulary would be random noise, and the resulting embeddings of the paragraphs would have high variance. Therefore, the tokenizer of the chosen retriever should ideally contain byte-level byte pair encoding (BBPE) \citep{wang2020neural}.

The third component of the RAG pipeline is the prompt. The prompt format to be used as the baseline only presents the paragraph, the question about the typological characteristic, clarifications regarding what the linguistic term refers to, and a closed set of answers, e.g., for WALS 81A “Dominant Order of Subject, Object and Verb”: “SVO”, “SOV”, “VOS”, “VSO”, “OSV”, “OVS”, “No dominant order”, and additionally “Not enough information” if the dominant word order in the language cannot be inferred from the grammar context.
The prompting strategy implemented in the pipeline includes the baseline prompt with additional description of the typological characteristic from WALS or Grambank with examples, as a variation of chain-of thought prompting \citep{wei2022chain}.

The last component of the RAG pipeline is the LLM. We use GPT-4o, OpenAI’s newest flagship model as of May 2024 with increased performance compared to GPT-4 \citep{achiam2023gpt}: its task is to determine the value of the feature, e.g., “4 cases” for WALS 49A – Number of Cases, based on the prompt and the paragraphs from the descriptive grammar.

In conclusion, the RAG pipeline for descriptive grammars can be called Retrieval Augmented Classification: compared to the more common applications of RAG, the task of the pipeline is to choose one of the values for a linguistic feature from a closed set instead of answering any forms of questions possible, including open-ended ones.

\section{The Benchmark for Rerankers}

\subsection{Data}

Passing the entire grammar to an LLM as a prompt in order to determine the value of a single typological feature is costly and computationally inefficient. Furthermore, the more crucial drawback of passing the unfiltered content to an LLM has been demonstrated in \citep{shi2023large}: quality of LLMs’ responses deteriorates on prompts containing irrelevant context.

Since state-of-the-art retrievers are also LLMs, it would be similarly computationally inefficient to pass the entire grammar to them in order to pass the resulting paragraphs to GPT-4o. Therefore state-of-the-art retrievers become rerankers (one of the advanced additions to the “naive” RAG pipeline): the 50 paragraphs chosen by BM25 are reranked by an LLM retriever, and the resulting top 20 paragraphs are inserted into the prompt for GPT-4o. 
 
The purpose of the benchmark for rerankers is evaluation of state-of-the-art retrievers/rerankers on WALS 81A: Order of Subject, Object and Verb. The benchmark contains 14 grammars written in English: two grammars from each of the six macroareas and two additional grammars for rare word orders (OVS, OSV). Each grammar was split into paragraphs, which were ranked by BM25 using the summary for the English Wikipedia article “Word order” as the query, as proposed in \citep{kornilov-2023-multilingual}.

The top 50 paragraphs with the highest ranks assigned by BM25 for each grammar were annotated according to the following principle:

\textbf{0} – the paragraph does not mention word order at all;

\textbf{1} – the paragraph mentions or describes word order in a construction other than the monotransitive construction (or order of morphemes/phonemes/clitics/etc.);

\textbf{2} – the paragraph mentions or describes the word order in the monotransitive construction (in a title of a section, in the table of contents, or in references);

\textbf{3} – the paragraph mentions or describes the word order in the monotransitive construction (in a paragraph in the main text);

\textbf{4} – the paragraph narrows down the word order in the monotransitive construction to several variants;

\textbf{5} – a linguist can unambiguously determine the constituent order in the monotransitive construction from the paragraph.

Examples of paragraphs for every relevance category are provided in Appendix~\ref{sec:relevance_categories}.

Among the resulting 700 paragraphs, 38.86\% (annotated with 0) are not relevant to order of any elements in a language. Furthermore, 42.43\% of the paragraphs (the ones annotated with 1, 2, 3) are potentially misleading data due to describing order of components in a language without explicitly stating the value of WALS 81A. Finally, 18.71\% of paragraphs provide relevant information; however, out of the relevant paragraphs, 31.29\% are only partially relevant (annotated with 4), and do not sufficiently elaborate on the constituent order in the monotransitive construction in order for the linguist to be able to determine the value of the feature. More detailed data on the benchmark is provided in Appendix~\ref{sec:rerankers_details}.

In conclusion, this dataset can be used as a benchmark for more advanced information retrieval methods, in order to evaluate their capabilities of filtering out noisy and potentially misleading data.

\subsection{Results: Evaluating Rerankers}

We use the benchmark presented in Section 4.1 to find the best performing retriever/reranker and incorporate it into the RAG pipeline.

As the metric for evaluating the rerankers, we chose NDCG@k (Normalized Discounted Cumulative Gain at k) \citep{jarvelin2002cumulated} over other metrics commonly used for evaluation of information retrieval systems: Recall@k, Mean Average Precision@k (MAP@k), and Mean Reciprocal Rank (MRR), since NDCG@k is the only metric among them that can take into account a scale of more than two relevant ranks: our scale contains six different categories of relevance (0-5) instead of a binary “1 = relevant, 0 = not relevant” distinction.

\begin{table*}
\small
\begin{tabular}{lllllllll}
\hline
\multicolumn{2}{l}{\multirow{3}{*}{}} & \multirow{3}{*}{\textbf{\makecell[l]{bge-en\\-icl}}} & \multirow{3}{*}{\textbf{\makecell[l]{stella\_\\en\_1.5B\\\_v5}}} & \multirow{3}{*}{\textbf{\makecell[l]{NV-Re-\\triever-\\v1}}} & \multirow{3}{*}{\textbf{\makecell[l]{gte-\\Qwen2-7B\\-instruct}}} & \multirow{3}{*}{\textbf{\makecell[l]{Linq-\\Embed-\\Mistral}}} & \multirow{3}{*}{\textbf{\makecell[l]{SFR-Em-\\bedding-\\2\_R}}} & \multirow{3}{*}{\textbf{\makecell[l]{SFR-Em-\\bedding-\\Mistral}}} \\
\multicolumn{2}{l}{} &  &  &  &  &  &  &  \\
\multicolumn{2}{l}{} &  &  &  &  &  &  &  \\
\hline
\multirow{2}{*}{\textbf{Default Instruct}} & \textbf{Term Only} & 0.7387 & 0.7503 & 0.6292 & 0.7598 & 0.7501 & 0.7739 & 0.7381 \\
 & \textbf{Wiki Summary} & 0.7261 & \textbf{0.7659} & \textbf{0.6693} & 0.7405 & 0.7465 & 0.7624 & 0.7527 \\
\multirow{2}{*}{\textbf{Specific Instruct}} & \textbf{Term Only} & 0.7075 & 0.7505 & 0.6492 & \textbf{0.7748} & \textbf{0.7753} & \textbf{0.7750} & 0.7625 \\
 & \textbf{Wiki Summary} & \textbf{0.7474} & 0.7601 & 0.6538 & 0.7520 & 0.7690 & 0.7676 &  \textbf{\underline{0.7776}} \\
 \hline
\end{tabular}
  \caption{\label{rerankers-benchmark-metrics}
    NDCG@20 scores on the reranker benchmark. \textit{Term Only} refers to using the sequence \textit{“Dominant word order (Order of Subject, Object, and Verb)”} as a query, while \textit{Wiki Summary} refers to using the Wikipedia summary from the page “Word order” in English as a query. The best result for each model is shown in \textbf{bold}, and the best result across all configuration variants is \underline{underlined}. The NDCG@20 score for BM25 without a reranker is \textbf{0.7494}.
  }
\end{table*}

\begin{table*}
\small
\begin{tabular}{llllll}
\hline
\multirow{3}{*}{} & \multirow{3}{*}{\textbf{\makecell[l]{MTEB\\Ranking}}} & \multirow{3}{*}{\textbf{\makecell[l]{Grammar\\Benchmark\\Ranking}}} & \multirow{3}{*}{\textbf{\makecell[l]{Best\\Performing\\Query}}} & \multirow{3}{*}{\textbf{\makecell[l]{Best\\Performing\\Instruct}}} & \multirow{3}{*}{\textbf{Parameters}} \\
\multicolumn{1}{l}{} &  &  &  &  \\
\multicolumn{1}{l}{} &  &  &  &  \\
\hline
\textbf{bge-en-icl} \citep{bge_embedding} & 1 & 6 & Term Only & Default & 7.11B \\
\textbf{stella\_en\_1.5B\_v5} & 2 & 5 & Wiki & Default & 1.54B \\
\textbf{NV-Retriever-v1} \citep{moreira2024nvretrieverimprovingtextembedding} & 3 & 7 & Wiki & Default & 7.11B \\
\textbf{gte-Qwen2-7B-instruct} \citep{li2023towards} & 4 & 4 & Term Only & Specific & 7.61B \\
\textbf{Linq-Embed-Mistral} \citep{LinqAIResearch2024} & 5 & 2 & Term Only & Specific & 7.11B \\
\textbf{SFR-Embedding-2\_R} \citep{SFR-embedding-2} & 6 & 3 & Term Only & Specific & 7.11B \\
\textbf{SFR-Embedding-Mistral} \citep{SFRAIResearch2024} & 10 & 1 & Wiki & Specific & 7.11B \\
\hline
\end{tabular}
  \caption{\label{rerankers-benchmark-rankings}
    Rankings on our benchmark, compared to MTEB, and the best performing configurations for the rerankers. The evaluation of the rerankers on our benchmark requires approximately 1.8 GPU hours in total on 1x NVIDIA A100 40GB.
  }
\end{table*}

The rerankers we chose are the 6 models with the best NDCG@10 score on the Massive Text Embedding Benchmark (MTEB) leaderboard \citep{muennighoff2023mtebmassivetextembedding} on the Retrieval task for English as of Aug 29, 2024, along with \textit{SFR-Embedding-Mistral} \citep{SFRAIResearch2024}, the top 1 model on MTEB as of May 19, 2024. The HuggingFace pages for the models are referenced in Appendix ~\ref{sec:huggingface_pages}.

Since all selected embedding models accept instructions, we tested two instruction options - Default Instruct, generic and commonly used for embedding models:

\begin{quote}
\textit{Given a web search query, retrieve relevant passages that answer the query}
\end{quote}

and Specific Instruct, tailored to our specific task:

\begin{quote}
\textit{Given a definition of a linguistic feature, retrieve relevant passages that let a linguist unambiguously determine the value of this feature in the described language}
\end{quote}

Apart from the two variants of the instruct, we use two variants of the query:  1. “\textit{Dominant word order (Order of Subject, Object, and Verb)}” and 2. the Wikipedia summary from the page “Word order” in English.

In addition to state-of-the-art embedding models, we evaluate BM25 itself as the baseline.

Plots with NDCG@k across all values of k for best performing configurations of
each reranker on all grammars are presented in Appendix~\ref{ndcg_all_grammars}, and the plots showcasing comparison between two query types are available in Appendix~\ref{wiki_vs_term_only}.

The results of reranker evaluation are presented in Table~\ref{rerankers-benchmark-metrics} and Table~\ref{rerankers-benchmark-rankings}. 

BM25, the baseline retriever, ranks 6th out of 8 and only marginally lags behind the top rerankers in terms of NDCG@20. It appears rational not to take \textit{NV-Retriever-V1}, which exhibits significantly lower performance compared to other rerankers, into account, since its sample code on HuggingFace yields different cosine similarity scores with different batch sizes (as of Aug 29, 2024); however, the other reranker outperformed by BM25 on the linguistic domain, \textit{bge-en-icl}, is top 1 on MTEB. Furthermore, the models' rankings on MTEB and the rankings on our benchmark have a strong negative correlation (Spearman's $\rho$ = -0.8571), indicating that the Information Retrieval subset of MTEB is not suitable for estimating embeddings' capabilities on the domain of linguistic descriptions.

In conclusion, based on the results on the benchmark described in this section, we have selected \textit{SFR-Embedding-Mistral} with Specific Instruct and Wikipedia Query as the reranker component for the RAG pipeline due to its superior NDCG@20 score compared to other models. Furthermore, the results on the benchmark for rerankers reinforce the decision to select BM25 as the base retriever, since despite being a rule-based method invented decades ago, it has proven itself to be on a similar level compared to state-of-the-art retrievers on the fragment of the linguistic domain presented in our benchmark. 

Following the selection of \textit{SFR-Embedding-Mistral} as the reranker, in the following section we proceed to describe the second benchmark created in order to assess the efficacy of the RAG pipeline as a comprehensive system.

\section{The Benchmark for RAG}

\subsection{Data}

The benchmark for the RAG pipeline comprises 148 descriptive grammars. We selected the ostensibly arbitrary number (initially 150) due to benchmarks with fewer than 100 items being unreliable as a tool of assessment: a wrong answer on one item would result in accuracy decreasing by more than one percentage point.

Selecting the grammars in a random fashion would defeat the purpose of the benchmark due to possible biases towards languages having the same descent (a particular family being overrepresented) or being spoken in the same area of the world, without regard to the factual proportions of languages spoken in different areas. Therefore, we used the Genus-Macroarea method described by \citep{miestamo2016sampling}: in particular, its implementation presented by \citep{cheveleva2023neutralization}. Identically to the method described in \citep{cheveleva2023neutralization}, we obtain the proportions of languages across macroareas from the list of genera from WALS, automatically choose descriptive grammars from Glottolog's \citep{glottolog} References\footnote{\url{https://glottolog.org/langdoc}} database, and create the sample anew, placing the limit of one language for each genus. Our sampling strategy differs from the one utilized by \citep{cheveleva2023neutralization} in that we limit our sample to grammars written in English, and instead of restricting references to \verb|grammar_sketch| and \verb|grammar| types from Glottolog, we allow references to contain other tags, as long as either \verb|grammar_sketch| or \verb|grammar| is present. The resulting proportions of languages across macroareas are presented in Table~\ref{macroarea-stratification}.

\begin{table}[ht]
\centering
\small
\begin{tabular}{ll}
\hline
\textbf{Macroarea} & \multicolumn{1}{l}{\textbf{Languages}} \\
\hline
Africa & 29  \\
Australia & 9  \\
Eurasia & 20  \\
North America & 25  \\
Papunesia & 39  \\
South America & 26 \\
\hline
\textbf{Total} & \textbf{148} \\
\hline
\end{tabular}
  \caption{\label{macroarea-stratification}
    Languages stratified by macroarea, adapted from \citep{miestamo2016sampling} and \citep{cheveleva2023neutralization}.
  }
\end{table}

We annotated four typological features for the languages described by the grammars in the benchmark.
The first feature is identical to the one presented in the benchmark for rerankers: WALS 81A – Order of Subject, Object, and Verb, as an example of a largely self-explanatory and straightforward feature, which is concentrated in one place in the majority of the grammars: if a paragraph mentions which basic constituent order the language has, the mention is in most cases explicit.

 The second annotated feature is from Grambank: GB 107 – “Can standard negation be marked by an affix, clitic or modification of the verb?” Despite being a binary feature, it cannot be reliably extracted by “naive” methods based on term frequency: in the cases when the author of the descriptive literature refers to the negation marker as a marker or a morpheme instead of explicitly calling it a clitic or an affix, the RAG system would have to rely on interlinear glosses and to distinguish between morphological and syntactic phenomena. Furthermore, even in the case when the negation marker is explicitly referred to as a clitic or an affix, it is necessary to determine from context if this marker is phonologically bound solely to the verb (which triggers the feature value = 1) or can be attached to any constituent in the clause (leaving the feature value at 0).
 
The third feature is a complex composed of seven binary features all related to polar (yes / no) questions; we have chosen it to evaluate the ability of LLMs to reason in the linguistic domain while taking into account several realizations of the same phenomenon simultaneously. This feature will be further referred to as WALS 116A*: despite being related to WALS 116A, it is more accurately described as an amalgamation of seven separate features from Grambank: GB257, GB260, GB262, GB263, GB264, GB286, and GB291.  This feature is essentially a multilabel classification with seven labels: each label/strategy from the set (Interrogative intonation, Interrogative word order, Clause-initial question particle, Clause-medial question particle, Clause-final question particle, Interrogative verb morphology, Tone) is annotated as 1 if it can be used to form polar questions in the described language, and as 0 otherwise.

The last feature is WALS 49A - Number of Cases. It has been chosen due to its quantitative nature, as opposed to binary features determining presence and absence of a particular phenomenon, and its scattered nature: the guideline for WALS 49A takes a liberal approach to determining the number of grammatical cases in a language, allowing adpositional clitics to be considered as case markers; consequently, the relevant information may be in entirely different sections of the grammar (nominal morphology for "traditional" case markers and syntax for adpositional clitics). Due to Number of Cases being the most time-consuming feature to annotate, we created a separate benchmark for it, consisting partially of grammars from the already existing benchmark and partially of grammars with existing annotations for WALS 49A, maintaining the macroarea proportions and adhering to the one-language-per-genus rule.

The distribution of values for each feature is presented in Appendix~\ref{sec:value_distrib}.

\subsection{Results: Evaluating the RAG Pipeline}

In order to evaluate any NLP task on a benchmark, it is crucial to run tests in order to determine if the LLM already possesses knowledge about the feature values in the benchmark. It is different from a contamination test: an LLM may possess knowledge about word order in a particular language from any resource, while a contamination test would determine if the LLM saw the grammar itself during pretraining. 

In order to set the baseline for GPT-4o, i.e., to estimate how well it performs on the linguistic domain without additional information from the grammar, we conducted a test on the RAG pipeline excluding the retrieval module. We prompted GPT-4o to determine the values of all benchmark features without the retrieved paragraphs: the only information given to the model consisted of the prompt and the Wikipedia summary for the article about the corresponding feature. The prompts are listed in Appendix~\ref{prompts}. For each feature, the baseline run was executed ten times in order to capture the variance of the results and to represent presence vs. absence of knowledge more accurately. It is crucial to note that the baseline was evaluated only on those languages where the grammar itself had sufficient information to identify the feature: 136, 146, 121, and 140 values out of 148 for WALS 81A, GB 107, WALS 116A, and WALS 49A respectively.

\begin{table*}[ht]
\centering
\small
\begin{tabular}{l|l|c|c|c|c|c}
 & \multicolumn{1}{c|}{\textbf{F1 mean}} & \textbf{Baseline} & \textbf{BM25} & \textbf{BM25+CoT} & \textbf{Reranker} & \textbf{Rer.+CoT} \\ \hline
 & \textbf{micro} & \cellcolor[HTML]{AFC9CF}0.5551 ± 0.0359 & \cellcolor[HTML]{8FB3BB}0.6892 & \cellcolor[HTML]{8CB1B9}0.7027 & \cellcolor[HTML]{8AB0B8}0.7095 & \cellcolor[HTML]{85ADB5}\textbf{0.7297} \\
 & \textbf{macro} & \cellcolor[HTML]{F1F6F7}0.2812 ± 0.0276 & \cellcolor[HTML]{88AEB7}\textbf{0.7179} & \cellcolor[HTML]{A2C0C7}0.6097 & \cellcolor[HTML]{A1BFC6}0.6141 & \cellcolor[HTML]{A1BFC6}0.6139 \\
\multirow{-3}{*}{\textbf{WALS 81A}} & \textbf{weighted} & \cellcolor[HTML]{B5CDD2}0.5328 ± 0.0337 & \cellcolor[HTML]{94B6BE}0.6694 & \cellcolor[HTML]{8FB3BB}0.6890 & \cellcolor[HTML]{91B5BD}0.6790 & \cellcolor[HTML]{8CB2BA}\textbf{0.6995} \\ \hline
 & \textbf{micro} & \cellcolor[HTML]{AAC6CC}0.5747 ± 0.0350 & \cellcolor[HTML]{95B8BF}0.6622 & \cellcolor[HTML]{97B9C0}0.6554 & \cellcolor[HTML]{92B5BD}0.6757 & \cellcolor[HTML]{8AB0B8}\textbf{0.7095} \\
 & \textbf{macro} & \cellcolor[HTML]{ABC6CC}0.5740 ± 0.0354 & \cellcolor[HTML]{ABC6CC}0.5718 & \cellcolor[HTML]{ABC6CC}0.5731 & \cellcolor[HTML]{A4C2C8}0.6011 & \cellcolor[HTML]{97B9C0}\textbf{0.6546} \\
\multirow{-3}{*}{\textbf{GB 107}} & \textbf{weighted} & \cellcolor[HTML]{ABC6CC}0.5724 ± 0.0361 & \cellcolor[HTML]{A5C2C9}0.5957 & \cellcolor[HTML]{A5C2C9}0.5959 & \cellcolor[HTML]{9FBEC5}0.6221 & \cellcolor[HTML]{93B6BE}\textbf{0.6713} \\ \hline
 & \textbf{micro} & \cellcolor[HTML]{D5E3E6}0.3986 ± 0.0238 & \cellcolor[HTML]{B8CFD4}0.5203 & \cellcolor[HTML]{B3CBD1}0.5405 & \cellcolor[HTML]{AFC9CF}\textbf{0.5541} & \cellcolor[HTML]{B9D0D5}0.5135 \\
 & \textbf{macro} & \cellcolor[HTML]{FFFFFF}0.2216 ± 0.0158 & \cellcolor[HTML]{C3D7DB}\textbf{0.4711} & \cellcolor[HTML]{C9DADE}0.4494 & \cellcolor[HTML]{CDDDE1}0.4332 & \cellcolor[HTML]{D4E2E5}0.4042 \\
\multirow{-3}{*}{\textbf{WALS 49A}} & \textbf{weighted} & \cellcolor[HTML]{E2EBED}0.3453 ± 0.0237 & \cellcolor[HTML]{B5CDD2}0.5314 & \cellcolor[HTML]{AFC9CF}0.5542 & \cellcolor[HTML]{AEC8CE}\textbf{0.5605} & \cellcolor[HTML]{B8CFD4}0.5185 \\ \hline
 & \textbf{micro} & \cellcolor[HTML]{C8DADE}0.4504 ± 0.0396 & \cellcolor[HTML]{699AA4}0.8446 & \cellcolor[HTML]{5C919C}\textbf{0.8986} & \cellcolor[HTML]{6B9BA5}0.8378 & \cellcolor[HTML]{61949F}0.8784 \\
 & \textbf{macro} & \cellcolor[HTML]{CFDEE2}0.4240 ± 0.0456 & \cellcolor[HTML]{6B9BA5}0.8392 & \cellcolor[HTML]{5D929D}\textbf{0.8946} & \cellcolor[HTML]{6C9CA6}0.8335 & \cellcolor[HTML]{6295A0}0.8739 \\
\multirow{-3}{*}{\textbf{\begin{tabular}[c]{@{}l@{}}interrog.\\ intonation \\ only\end{tabular}}} & \textbf{weighted} & \cellcolor[HTML]{D3E1E4}0.4068 ± 0.0484 & \cellcolor[HTML]{6999A4}0.8480 & \cellcolor[HTML]{5C919C}\textbf{0.9007} & \cellcolor[HTML]{6A9AA5}0.8415 & \cellcolor[HTML]{61949F}0.8810 \\ \hline
 & \textbf{micro} & \cellcolor[HTML]{4C8693}0.9653 ± 0.0076 & \cellcolor[HTML]{46828F}\textbf{0.9932} & \cellcolor[HTML]{478390}0.9865 & \cellcolor[HTML]{478390}0.9865 & \cellcolor[HTML]{498491}0.9797 \\
 & \textbf{macro} & \cellcolor[HTML]{95B7BF}0.6641 ± 0.0460 & \cellcolor[HTML]{528A96}\textbf{0.9427} & \cellcolor[HTML]{5D919D}0.8965 & \cellcolor[HTML]{6395A0}0.8715 & \cellcolor[HTML]{6D9DA7}0.8281 \\
\multirow{-3}{*}{\textbf{\begin{tabular}[c]{@{}l@{}}interrog. \\ word \\ order\end{tabular}}} & \textbf{weighted} & \cellcolor[HTML]{4D8793}0.9611 ± 0.0064 & \cellcolor[HTML]{45818E}\textbf{0.9936} & \cellcolor[HTML]{47828F}0.9878 & \cellcolor[HTML]{478390}0.9865 & \cellcolor[HTML]{498490}0.9808 \\ \hline
 & \textbf{micro} & \cellcolor[HTML]{86ADB6}0.7264 ± 0.0402 & \cellcolor[HTML]{5B909B}0.9054 & \cellcolor[HTML]{578E99}0.9189 & \cellcolor[HTML]{518995}\textbf{0.9459} & \cellcolor[HTML]{568D98}0.9257 \\
 & \textbf{macro} & \cellcolor[HTML]{B2CBD0}0.5446 ± 0.0502 & \cellcolor[HTML]{75A1AB}0.7977 & \cellcolor[HTML]{6C9CA6}0.8335 & \cellcolor[HTML]{5F939E}\textbf{0.8890} & \cellcolor[HTML]{6899A3}0.8504 \\
\multirow{-3}{*}{\textbf{\begin{tabular}[c]{@{}l@{}}clause-\\ initial \\ particle\end{tabular}}} & \textbf{weighted} & \cellcolor[HTML]{83ABB4}0.7371 ± 0.0337 & \cellcolor[HTML]{5B909B}0.9054 & \cellcolor[HTML]{578D99}0.9205 & \cellcolor[HTML]{518995}\textbf{0.9470} & \cellcolor[HTML]{558C98}0.9278 \\ \hline
 & \textbf{micro} & \cellcolor[HTML]{BFD4D8}0.4909 ± 0.0409 & \cellcolor[HTML]{87AEB6}0.7230 & \cellcolor[HTML]{7FA8B1}0.7568 & \cellcolor[HTML]{8CB1B9}0.7027 & \cellcolor[HTML]{78A4AD}\textbf{0.7838} \\
 & \textbf{macro} & \cellcolor[HTML]{C1D5DA}0.4808 ± 0.0415 & \cellcolor[HTML]{88AFB7}0.7173 & \cellcolor[HTML]{81AAB3}0.7463 & \cellcolor[HTML]{8DB2BA}0.6972 & \cellcolor[HTML]{7BA6AF}\textbf{0.7717} \\
\multirow{-3}{*}{\textbf{\begin{tabular}[c]{@{}l@{}}clause-\\ final \\ particle\end{tabular}}} & \textbf{weighted} & \cellcolor[HTML]{C5D8DC}0.4661 ± 0.0428 & \cellcolor[HTML]{85ACB5}0.7314 & \cellcolor[HTML]{7DA7B0}0.7644 & \cellcolor[HTML]{89B0B8}0.7116 & \cellcolor[HTML]{77A3AC}\textbf{0.7902} \\ \hline
 & \textbf{micro} & \cellcolor[HTML]{92B5BD}0.6760 ± 0.0367 & \cellcolor[HTML]{6D9CA6}0.8311 & \cellcolor[HTML]{60939E}\textbf{0.8851} & \cellcolor[HTML]{6B9BA5}0.8378 & \cellcolor[HTML]{6698A2}0.8581 \\
 & \textbf{macro} & \cellcolor[HTML]{BCD2D7}0.5011 ± 0.0544 & \cellcolor[HTML]{7DA7B0}0.7638 & \cellcolor[HTML]{6F9EA8}\textbf{0.8201} & \cellcolor[HTML]{7BA6AF}0.7709 & \cellcolor[HTML]{78A4AD}0.7832 \\
\multirow{-3}{*}{\textbf{\begin{tabular}[c]{@{}l@{}}clause-\\ medial  \\ particle\end{tabular}}} & \textbf{weighted} & \cellcolor[HTML]{95B7BF}0.6644 ± 0.0355 & \cellcolor[HTML]{6A9AA4}0.8439 & \cellcolor[HTML]{5F939E}\textbf{0.8888} & \cellcolor[HTML]{6899A4}0.8496 & \cellcolor[HTML]{6597A1}0.8641 \\ \hline
 & \textbf{micro} & \cellcolor[HTML]{7CA6B0}0.7678 ± 0.0185 & \cellcolor[HTML]{75A2AB}0.7973 & \cellcolor[HTML]{6D9CA6}\textbf{0.8311} & \cellcolor[HTML]{6E9DA7}0.8243 & \cellcolor[HTML]{6D9CA6}\textbf{0.8311} \\
 & \textbf{macro} & \cellcolor[HTML]{C8DADE}0.4533 ± 0.0311 & \cellcolor[HTML]{8AB0B8}0.7075 & \cellcolor[HTML]{84ACB4}0.7354 & \cellcolor[HTML]{85ADB5}0.7282 & \cellcolor[HTML]{82ABB3}\textbf{0.7419} \\
\multirow{-3}{*}{\textbf{\begin{tabular}[c]{@{}l@{}}interrog.\\ verb\\ morphology\end{tabular}}} & \textbf{weighted} & \cellcolor[HTML]{8AB0B8}0.7102 ± 0.0160 & \cellcolor[HTML]{709EA8}0.8192 & \cellcolor[HTML]{699AA4}0.8451 & \cellcolor[HTML]{6B9BA5}0.8396 & \cellcolor[HTML]{699AA4}\textbf{0.8465} \\ \hline
 & \textbf{micro} & \cellcolor[HTML]{558C98}0.9273 ± 0.0128 & \cellcolor[HTML]{568D98}0.9257 & \cellcolor[HTML]{4E8793}0.9595 & \cellcolor[HTML]{4C8693}0.9662 & \cellcolor[HTML]{4A8592}\textbf{0.9730} \\
 & \textbf{macro} & \cellcolor[HTML]{B9D0D5}0.5147 ± 0.0573 & \cellcolor[HTML]{82ABB4}0.7407 & \cellcolor[HTML]{6F9DA8}0.8225 & \cellcolor[HTML]{6697A2}\textbf{0.8594} & \cellcolor[HTML]{6899A4}0.8501 \\
\multirow{-3}{*}{\textbf{tone}} & \textbf{weighted} & \cellcolor[HTML]{5A8F9B}0.9104 ± 0.0121 & \cellcolor[HTML]{538A96}0.9390 & \cellcolor[HTML]{4D8693}0.9637 & \cellcolor[HTML]{4B8592}0.9704 & \cellcolor[HTML]{4A8592}\textbf{0.9730}
\end{tabular}
  \caption{\label{rag_results}
   F1 scores across all configurations of the pipeline. All five configurations contain prompts from Appendix~\ref{prompts} with Wikipedia summaries for corresponding features. The \textit{Baseline} column refers to prompting GPT-4o without materials from grammars (i.e, without RAG). \textit{BM25} refers to using 50 paragraphs retrieved by BM25. \textit{Reranker} refers to using 20 paragraphs selected from these 50 by \textit{SFR-Embedding-Mistral}. \textit{CoT} refers to Chain-of-Thought (adding instructions and examples from WALS or Grambank to the prompt). The best F1 value in each row is shown in \textbf{bold}. The values after the ± sign in the \textit{Baseline} column are the sample standard deviation across 10 runs. Each of the remaining four configurations was executed only once.
  }
\end{table*}

We subsequently integrated the retriever/ reranker component into the pipeline and tested GPT-4o on four prompt configurations: two options for the retrieved information (50 paragraphs from BM25 / 20 paragraphs from a reranker on top of BM25) and two options for prompting (default / with Chain-of-Thought). The Chain-of-Thought prompts are the default RAG prompts concatenated with guidelines and examples from corresponding chapters in WALS and Grambank. We deemed Grambank chapters particularly suitable for the purpose of Chain-of-Thought prompting, because each chapter comprises: the summary of the feature with the clarifications on ambiguous linguistic terms
(i.e., it is explicitly stated in GB263 that only neutral polar questions should be considered in its context, while leading polar questions should be ignored); the step-by-step algorithm intended to instruct human annotators on determining the value of the feature; and examples from the world’s languages with interlinear glosses and explanations of the reason why the feature is present (or missing) in the language.

More details regarding the prompt configurations are available in Appendix~\ref{sec:prompt_details}.

The temperature value we chose for GPT-4o is 0.2, since our tasks requires the model to incline towards more deterministic behavior.

Running the rerankers on the RAG benchmark for all configurations requires approximately 7.2 GPU hours in total on 1x NVIDIA A100 40GB.

The results for all RAG configurations are presented in Table~\ref{rag_results}. All RAG configurations outperform the baselines. The observation that macro-averaged F1 scores tend to be higher than micro-averaged F1 scores—which equate to accuracy, given that we do not treat any features as multilabel classifications—suggests that the RAG pipeline is more effective with more frequent classes, struggling to address the class imbalance present in the typological profiles of the world's languages. The results are overall inconsistent, and it is important to note that the Chain-of-Thought approach does not always improve upon its default counterpart, contrary to the expectation for Chain-of-Thought to excel on intensive reasoning tasks.

An additional ablation study is provided in Appendix~\ref{sec:ablation}.

\section{Discussion}
Expanding the MTOB approach could significantly benefit from standardizing descriptive grammars of various languages into a uniform format, leveraging databases such as Grambank or WALS.
However, we can state that in the non-contaminated environment demonstrated in this work the descriptive linguistic texts still pose a significant challenge.
While generally machine reading can be perceived as a “solved task”, the results on linguistic features show that descriptive grammars remain a non-saturated material showcasing LLMs' weak spots.

\section{Conclusion}
In this paper, we introduced two benchmarks for evaluation of methods that combine Retrieval Augmented Generation with large language models to extract and classify typological features from descriptive grammars. 

We also provided an open-source pipeline for linguistic information extraction, which has significant potential to improve NLP applications for under-resourced languages.

The proposed pipeline, alongside the presented benchmarks, revealed that BM25, a language-agnostic information retrieval method, is comparable in quality to state-of-the-art embedding-based methods on the task of retrieving information from descriptive grammars, and can be used as a RAG component on the linguistic domain. Furthermore, despite the notion that the machine reading task has mostly been resolved, the complexities inherent in linguistic texts still present challenges. While the advancements in language models have made significant strides in handling various types of texts, the results obtained on our benchmarks suggest that it remains premature to assert their effectiveness on the domain of linguistic descriptions.

Our contributions lay the groundwork for extending the capabilities of LLMs to handle complex linguistic data, such as grammatical descriptions. This work represents a crucial step toward better supporting low-resource languages in NLP. Future work could further optimize the retrieval and classification processes, expand the benchmark to include more languages, and explore practical applications of information extraction on the linguistic domain, such as cross-lingual typological analysis and machine translation for extremely low-resource languages.

\section{Limitations}

One of the limitations of our work is the number of languages presented: despite efforts to choose a stratification method that would yield a representative sample, 148 languages (the chosen number for each feature) constitute only about 2\% of the world's languages.

Another limitation is potential presence of erroneous data. WALS and Grambank currently list 2,662 and 2,467 languages respectively, which
falls into the range of 30-40\% of the world’s known languages (fewer than 60\% with grammars available). Furthermore, most language profiles are not full: there are only 5 and 841 languages annotated for the least “popular” feature in WALS and Grambank respectively. Additionally, \citep{baylor2023past} demonstrate that typological databases contain erroneous entries and discrepancies, reporting 69.04\% average agreement between WALS and Grambank across six typological features (agreement score for each feature obtained by averaging scores for macroareas). A portion of the discrepancies is most likely attributable to human error. Despite extensive efforts to eliminate the errors and fill the gaps, our benchmark is not to be considered perfect data either, since the first author was the only annotator.

Furthermore, due to high computational costs of running RAG on GPT-4o with extensive excerpts from grammars, we limited our experiments (excluding the baseline, which was executed 10 times) to running every RAG configuration only once. 

Finally, one of the most crucial limitations of this paper is the inability to make the benchmark for the RAG pipeline fully open-source, since the majority of grammars in the benchmark are under copyright.

\section{Ethical Statement}
In pursuing the advancement of machine translation and natural language processing through the use of descriptive grammars, we must address critical ethical concerns, particularly relating to copyright. Most grammars in the benchmark have copyright licenses which prohibit their reproduction fully. 

The benchmark for rerankers and the benchmark for RAG differ in the following way: in the benchmark for rerankers, each individual paragraph is annotated according to relevance to WALS 81A (to be used for retrieval assessment), whereas the benchmark for RAG does not have annotations for individual paragraphs. Therefore, we release our codebase and materials under MIT and publish the first benchmark fully, since it can be considered as a derivative work, and its publication for research purposes falls under fair use.

To ensure compliance with copyright laws and respect for intellectual property rights, we do not publish the second benchmark fully; we only publish the list of grammars and the list of page numbers where the relevant information for each feature can be found.

Furthermore, we advocate for the development of collaborative agreements with copyright holders
in order to obtain permission for making at least some of the full texts of the grammars available
as open-source. We believe that addressing these copyright issues is crucial for the sustainable and ethical advancement of language technologies.


\bibliography{custom}

\clearpage

\appendix

\section{Rerankers: HuggingFace}
\label{sec:huggingface_pages}

\begin{enumerate}
  \item bge-en-icl \citep{bge_embedding}: \url{https://huggingface.co/BAAI/bge-en-icl} \\ License: Apache 2.0
  \item stella\_en\_1.5B\_v5: \url{https://huggingface.co/dunzhang/stella_en_1.5B_v5} \\ License: MIT
  \item NV-Retriever-v1 \citep{moreira2024nvretrieverimprovingtextembedding}: \url{https://huggingface.co/nvidia/NV-Retriever-v1} \\ License: NVIDIA license agreement
  \item gte-Qwen2-7B-instruct \citep{li2023towards}: \url{https://huggingface.co/Alibaba-NLP/gte-Qwen2-7B-instruct} \\ License: Apache 2.0
  \item Linq-Embed-Mistral \citep{LinqAIResearch2024}: \url{https://huggingface.co/Linq-AI-Research/Linq-Embed-Mistral} \\ License: CC-BY-NC-4.0
  \item SFR-Embedding-2\_R \citep{SFR-embedding-2}: \url{https://huggingface.co/Salesforce/SFR-Embedding-2_R} \\ License: CC-BY-NC-4.0
  \item SFR-Embedding-Mistral \citep{SFRAIResearch2024}: \url{https://huggingface.co/Salesforce/SFR-Embedding-Mistral} \\ License: CC-BY-NC-4.0
\end{enumerate}

\section{Benchmark for Rerankers: Examples for Relevance Categories}
\label{sec:relevance_categories}

\textbf{0} — the paragraph does not mention word order at all.

\begin{quote}
\textit{In order to express ‘from’, these demonstrative members must take the abla­tive-1 suffix (-ngomay), like all other adverb and also nouns, e.g. yarro-ngomay ‘here-ABL1’ in (3-134), (4-13), (4-18-b), (4-77), (4-114).} \citep[p. 181]{tsunoda2011grammar}
\end{quote}

Annotated with 0: the paragraph only describes demonstratives without any mentions of orders of elements in the language.

Paragraphs that contain examples with glosses without explicit mentions of word order were similarly annotated with 0: examples without context should not be treated as evidence of a language having a particular word order, because a language may have no dominant word order.

\textbf{1} — the paragraph mentions or describes word order in a construction other than the monotransitive construction (or order of morphemes/phonemes/clitics/etc.), since WALS 81A refers to the word order in the monotransitive construction with the verb in the declarative mood in particular.

\begin{quote}
\textit{Table 36 shows that propositional enclitics are ordered in relation to
   each other. The directional enclitics -(e)nhdhi 'TOWARDS' and -(e)ya
   'AWAY' are mutually exclusive.} \citep[p. 269]{ford1998description}
\end{quote}

Annotated with 1: the paragraph describes order of enclitics instead of the constituent order in the monotransitive construction.

\textbf{2} — the paragraph mentions or describes the word order in the monotransitive construction (in a title of a section, in the table of contents, or in references).

\begin{quote}
\textit{ 736
   27.3 Word order at the clause level . . . . . . . . . . . . . . . . . .  } \citep[p. xxiv]{forker2013grammar}
\end{quote}

Annotated with 2: this chunk is a fragment of a table of contents.

\textbf{3} — the paragraph mentions or describes the word order in the monotransitive construction (in a paragraph in the main text). 

\begin{quote}
\textit{Alsea, Siuslaw, and Coos have been tentatively categorized as having VOS as their basic word order, by Greenberg (1966), on the basis of the fact that Greenberg found VOS to be the most common order of subject, object, and verb in these languages.} \citep[p. 482]{morgan1991description}
\end{quote}

Annotated with 3: the grammar describes the language Kutenai, but this paragraph mentions the constituent order in the monotransitive construction in other languages.

\textbf{4} — the paragraph narrows down the word order in the monotransitive construction to several variants.

\begin{quote}
\textit{As a consequence of its predominant verb-medial order, Qaqet does not have any clause chaining and/or switch reference <...>} \citep[p. 19]{hellwig2019grammar}
\end{quote}

Annotated with 4: the mention that Qaqet has a predominantly verb-medial order narrows the seven logically possible variants to “SVO”, “OVS”, and “No dominant order”.

\textbf{5} — a linguist can unambiguously determine the constituent order in the monotransitive construction from the paragraph.

\begin{quote}
\textit{<...> The constituent order in relative clauses is SOV, as in main clauses. The subject in relative clauses is obligatorily encoded as genitive, while all other constituents appear as they would in an independent verbal clause.}
\citep[p. 254]{wegener2012grammar}
\end{quote}

Annotated with 5: the word order is explicitly mentioned in the paragraph.

\begin{quote}
\textit{One might expect that the peculiar constituent order of Urarina would also be subject to pressure from Spanish (a notorious A V O / S V language), but significant changes to constituent order in Urarina are not observed. As mentioned in §18.3, there are a few isolated examples of an S or A argument occurring in preverbal position that cannot be accounted for in terms of the predicted features (focus, emphasis, negation). Beside that, in one of the dialects investigated further above (Copal), two examples with an O argument in postverbal position were observed. While such examples are extremely rare, one could of course attribute these to the influence of Spanish.} \citep[p. 899]{olawsky2006grammar}
\end{quote}

Annotated with 5. Although there is no explicit mention of the word order in Urarina, it is described in the paragraph that there are only isolated examples of the subject argument in Urarina occurring in the preverbal position and of the object argument occurring in the postverbal position. Consequently. the only possible logical variant that is possible for Urarina is OVS, contrary to the immediately obvious mention of SVO (A V O) in Spanish. Extracting information from such paragraphs based solely on term frequency would be suboptimal.

\begin{table*}
\section{Benchmark for Rerankers: Details}
\label{sec:rerankers_details}
\includegraphics[width=0.10\linewidth]{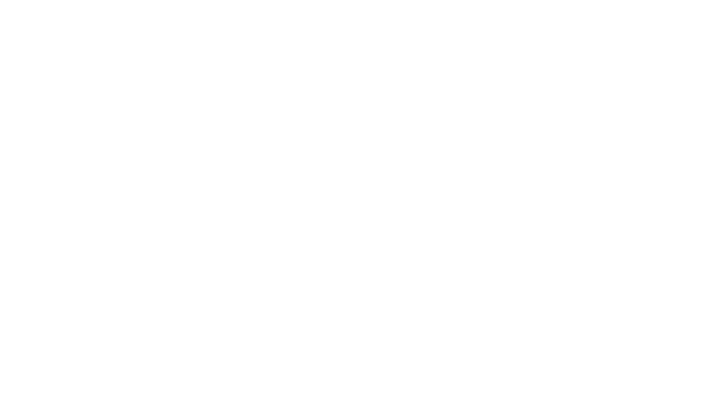}
\centering
\begin{tabular}{lllll}
\hline
{\textbf{Grammar}} & {\textbf{Language}} & {\textbf{Family}} & {\textbf{Macroarea}} & {\textbf{Word order}} \\
 \hline
\citep{campbell2017grammar} & Gã & Niger-Congo & Africa & SVO \\
\citep{Newman2002TheHL} & Hausa & Afro-Asiatic & Africa & No mention \\
\citep{georg2007descriptive} & Ket & Yeniseian & Eurasia & No mention \\
\citep{forker2013grammar} & Hinuq & Nakh-Daghestanian & Eurasia & SOV \\
\citep{ford1998description} & Emmi & Western Daly & Australia & No dom. order \\
\citep{tsunoda2011grammar} & Warrongo & Pama-Nyungan & Australia & SOV \\
\citep{morgan1991description} & Kutenai & Kutenai & North America & VOS \\
\citep{dunn1979reference} & Tsimshian (Coast) & Tsimshianic & North America & VSO \\
\citep{elliott2021grammar} & Enxet Sur & Lengua & South America & No dom. order \\
\citep{sakel2011grammar} & Mosetén & Mosetenan & South America & SVO \\
\citep{hellwig2019grammar} & Qaqet & Baining & Papunesia & SVO \\
\citep{wegener2012grammar} & Savosavo & Solomons East Papuan & Papunesia & SOV \\
\citep{olawsky2006grammar} & Urarina & Urarina & South America & OVS \\
\citep{weir1986footprints} & Nadëb & Nadahup & South America & OSV \\
\hline
\end{tabular}
  \caption{\label{retrievers-benchmark}
    Languages featured in the benchmark for rerankers. “No mention” in the column “Word order” indicates that the 50 paragraphs chosen by BM25 for this grammar do not allow to narrow down the word order in the language to one feature.
  }
\end{table*}

\begin{table*}
\begin{tabular}{llllllllll}
\hline
 & \multicolumn{6}{l}{\textbf{Annotations}} &  &  &  \\

\multirow{-2}{*}{\textbf{Grammar}} & \textbf{0} & \textbf{1} & \textbf{2} & \textbf{3} & \textbf{4} & \textbf{5} & \multirow{-2}{*}{\textbf{\makecell[l]{0 \\ (irrelevant)}}} & \multirow{-2}{*}{\textbf{\makecell[l]{1-3 \\ (misleading)}}} & \multirow{-2}{*}{\textbf{\makecell[l]{4-5 \\ (relevant)}}} \\
\hline
\citep{campbell2017grammar} & 34 & 7 & 0 & 2 & 0 & 7 & 0.68 & 0.18 & 0.14 \\
\citep{Newman2002TheHL} & 15 & 30 & 3 & 1 & 1 & 0 & 0.30 & 0.68 & 0.02 \\
\citep{georg2007descriptive} & 36 & 13 & 0 & 0 & 1 & 0 & 0.72 & 0.26 & 0.02 \\
\citep{forker2013grammar} & 9 & 14 & 10 & 6 & 4 & 7 & 0.18 & 0.60 & 0.22 \\
\citep{ford1998description} & 28 & 19 & 0 & 0 & 0 & 3 & 0.56 & 0.38 & 0.06 \\
\citep{tsunoda2011grammar} & 8 & 23 & 7 & 5 & 3 & 4 & 0.16 & 0.70 & 0.14 \\
\citep{morgan1991description} & 0 & 12 & 10 & 11 & 4 & 13 & 0.00 & 0.66 & 0.34 \\
\citep{dunn1979reference} & 45 & 0 & 1 & 2 & 0 & 2 & 0.90 & 0.06 & 0.04 \\
\citep{elliott2021grammar} & 5 & 18 & 2 & 4 & 3 & 18 & 0.10 & 0.48 & 0.42 \\
\citep{sakel2011grammar} & 11 & 19 & 5 & 6 & 4 & 5 & 0.22 & 0.60 & 0.18 \\
\citep{hellwig2019grammar} & 24 & 10 & 1 & 0 & 7 & 8 & 0.48 & 0.22 & 0.30 \\
\citep{wegener2012grammar} & 17 & 16 & 7 & 3 & 1 & 6 & 0.34 & 0.52 & 0.14 \\
\citep{olawsky2006grammar} & 6 & 17 & 3 & 2 & 8 & 14 & 0.12 & 0.44 & 0.44 \\
\citep{weir1986footprints} & 34 & 4 & 2 & 2 & 5 & 3 & 0.68 & 0.16 & 0.16 \\
\hline
\textbf{Total} & \textbf{272} & \textbf{202} & \textbf{51} & \textbf{44} & \textbf{41} & \textbf{90} & \textbf{0.3886} & \textbf{0.4243} & \textbf{0.1871} \\
\hline
\end{tabular}
  \caption{\label{retrievers-benchmark-annotation}
    Distribution of annotation categories in the benchmark for rerankers.
  }
\end{table*}

\begin{figure*}
\section{Benchmark for Rerankers: Dynamics for NDCG@k}
\label{ndcg_all_grammars}
    \centering
    \begin{minipage}[t]{0.48\linewidth}
        \scalebox{0.33}{\input{campbell2017grammar.pgf}}
    \end{minipage}
    \begin{minipage}[t]{0.48\linewidth}
        \scalebox{0.33}{\input{dunn1979reference.pgf}}
    \end{minipage}
    \begin{minipage}[t]{0.48\linewidth}
        \scalebox{0.33}{\input{elliott2021grammar.pgf}}
    \end{minipage}
    \begin{minipage}[t]{0.48\linewidth}
        \scalebox{0.33}{\input{ford1998description.pgf}}
    \end{minipage}
    \begin{minipage}[t]{0.48\linewidth}
        \scalebox{0.33}{\input{forker2013grammar.pgf}}
    \end{minipage}
    \begin{minipage}[t]{0.48\linewidth}
        \scalebox{0.33}{\input{georg2007descriptive.pgf}}
    \end{minipage}
\end{figure*}

\begin{figure*}
    \centering
    \begin{minipage}[t]{0.48\linewidth}
        \scalebox{0.33}{\input{hellwig2019grammar.pgf}}
    \end{minipage}
    \begin{minipage}[t]{0.48\linewidth}
        \scalebox{0.33}{\input{morgan1991description.pgf}}
    \end{minipage}
    \begin{minipage}[t]{0.48\linewidth}
        \scalebox{0.33}{\input{newman2000hausa.pgf}}
    \end{minipage}
    \begin{minipage}[t]{0.48\linewidth}
        \scalebox{0.33}{\input{olawsky2011grammar.pgf}}
    \end{minipage}
    \begin{minipage}[t]{0.48\linewidth}
        \scalebox{0.33}{\input{sakel2011grammar.pgf}}
    \end{minipage}
    \begin{minipage}[t]{0.48\linewidth}
        \scalebox{0.33}{\input{tsunoda2011grammar.pgf}}
    \end{minipage}
\end{figure*}

\begin{figure*}[t!]
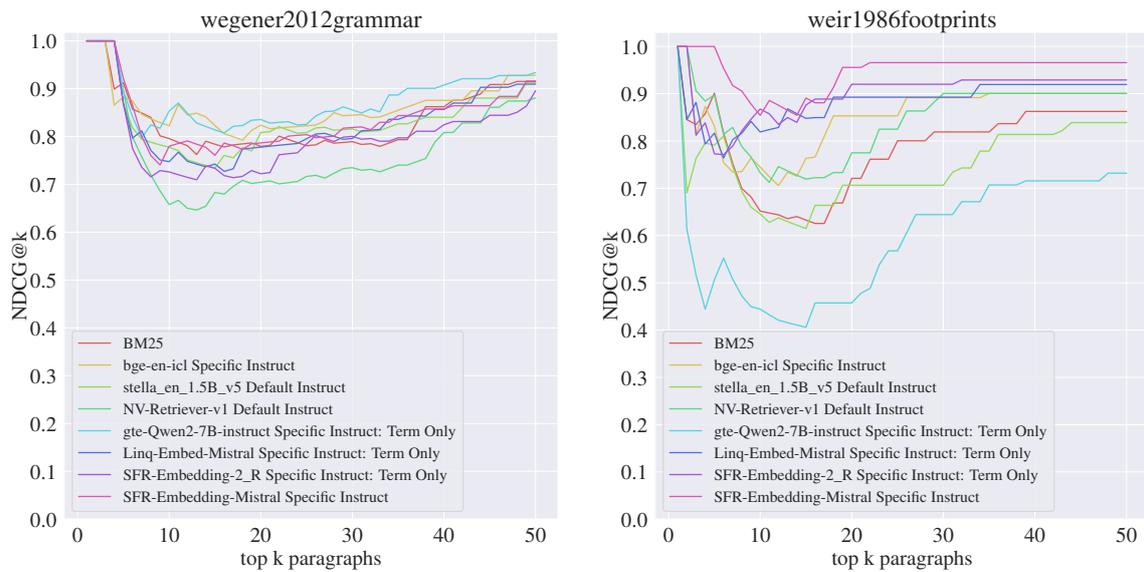

    \centering
    \begin{minipage}[t]{0.48\linewidth}
        \scalebox{0.33}{\input{wegener2012grammar.pgf}}
    \end{minipage}
    \begin{minipage}[t]{0.48\linewidth}
        \scalebox{0.33}{\input{weir1986footprints.pgf}}
    \end{minipage}
    \caption{\label{ndcg_all_grammars_table}
    NDCG@k across all values of k for best performing configurations for each model.
    }
\includegraphics[width=0.10\linewidth]{empty.png}

\begin{flushleft}

    One of the most evident phenomena indicated by the plots is low NDCG@k scores at small values of k on \citep{Newman2002TheHL}, indicating that all models ranked an irrelevant paragraph first. The “irrelevant” paragraph in question appears to be the following:

    \begin{quote}
    \textit{WORD ORDER The basic word order in sentences with i.o.’s is V + i.o.+ (d.o.),i.e., verb followedimmediately by the sh. indirect object followed by the direct object (if present). This word order is the same whether the i.0.is a noun or a pronoun, e.g., ura v3amigrate Misa ya kawo [wa tsohuwa];9, ruwaMusa brought the old woman water.} \citep{Newman2002TheHL}
    \end{quote}
    
    This paragraph describes the basic word order in a ditransitive construction and has been annotated with 1 (“the paragraph mentions or describes word order in a construction other than the monotransitive construction”). While a linguist may make a rational assumption that the monotransitive construction also follows the VO pattern, it would be incorrect to make assumptions in most similar cases: for instance, WALS lists 13 languages that have the SVO dominant order for transitive constructions, but VS for intransitive constructions, and any newly discovered language may potentially violate a principle previously considered a universal.
\end{flushleft}

\includegraphics[width=0.90\linewidth]{empty.png}

\end{figure*}

\begin{figure*}
\section{Benchmark for Rerankers: Wikipedia Summary vs. Term Only}
\label{wiki_vs_term_only}
    \centering
    \begin{minipage}[t]{0.48\linewidth}
        \scalebox{0.33}{\input{bge-en-icl_Default_Instruct.pgf}}
    \end{minipage}
    \begin{minipage}[t]{0.48\linewidth}
        \scalebox{0.33}{\input{bge-en-icl_Specific_Instruct.pgf}}
    \end{minipage}
    \begin{minipage}[t]{0.48\linewidth}
        \scalebox{0.33}{\input{stella_en_1.5B_v5_Default_Instruct.pgf}}
    \end{minipage}
    \begin{minipage}[t]{0.48\linewidth}
        \scalebox{0.33}{\input{stella_en_1.5B_v5_Specific_Instruct.pgf}}
    \end{minipage}
    \begin{minipage}[t]{0.48\linewidth}
        \scalebox{0.33}{\input{NV-Retriever-v1_Default_Instruct.pgf}}
    \end{minipage}
    \begin{minipage}[t]{0.48\linewidth}
        \scalebox{0.33}{\input{NV-Retriever-v1_Specific_Instruct.pgf}}
    \end{minipage}
\end{figure*}

\begin{figure*}
    \centering
    \begin{minipage}[t]{0.48\linewidth}
        \scalebox{0.33}{\input{gte-Qwen2-7B-instruct_Default_Instruct.pgf}}
    \end{minipage}
    \begin{minipage}[t]{0.48\linewidth}
        \scalebox{0.33}{\input{gte-Qwen2-7B-instruct_Specific_Instruct.pgf}}
    \end{minipage}
    \begin{minipage}[t]{0.48\linewidth}
        \scalebox{0.33}{\input{Linq-Embed-Mistral_Default_Instruct.pgf}}
    \end{minipage}
    \begin{minipage}[t]{0.48\linewidth}
        \scalebox{0.33}{\input{Linq-Embed-Mistral_Specific_Instruct.pgf}}
    \end{minipage}
    \begin{minipage}[t]{0.48\linewidth}
        \scalebox{0.33}{\input{SFR-Embedding-2_R_Default_Instruct.pgf}}
    \end{minipage}
    \begin{minipage}[t]{0.48\linewidth}
        \scalebox{0.33}{\input{SFR-Embedding-2_R_Specific_Instruct.pgf}}
    \end{minipage}
\end{figure*}

\begin{figure*}
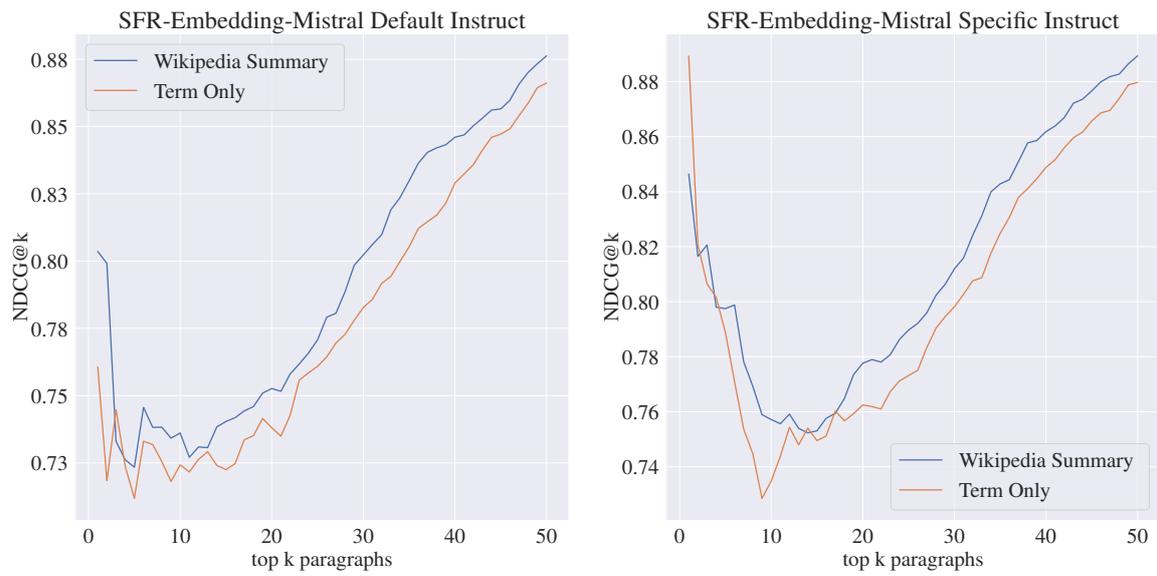

    \centering
    \begin{minipage}[t]{0.48\linewidth}
        \scalebox{0.33}{\input{SFR-Embedding-Mistral_Default_Instruct.pgf}}
    \end{minipage}
    \begin{minipage}[t]{0.48\linewidth}
        \scalebox{0.33}{\input{SFR-Embedding-Mistral_Specific_Instruct.pgf}}
    \end{minipage}
    \caption{\label{wiki_vs_term_only_table}
    Mean NDCG@k for all grammars: Wikipedia Summary vs Term Only.
    }

\includegraphics[width=1.70\linewidth]{empty.png}

\end{figure*}

\begin{figure*}
\section{Benchmark for RAG: Prompts}
\label{prompts}

\includegraphics[width=0.30\linewidth]{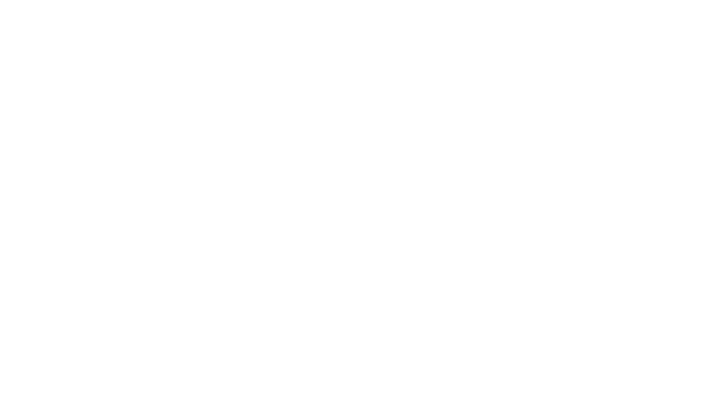}

Prompt for Word Order (partially based on WALS 81A):

\begin{quote}
\textit{Please determine the dominant word order (order of subject, object, and verb) in the language <...>.\\
The term "dominant word order" in the context of this feature refers to the dominant order of constituents in declarative sentences, in the case where both the subject and the object participants are nouns.\\
Reply with one of the 7 following options: SOV, SVO, VOS, VSO, OVS, OSV, No dominant order.\\
1. Provide the reasoning for the chosen option.\\
2. After the reasoning, output the word "Conclusion:" and the chosen option at the end of your response.}
\end{quote}

\end{figure*}

\begin{figure*}

Prompt for Standard Negation (partially based on GB107):

\begin{quote}
\textit{Please determine if standard negation in the language <...> can be marked by a modification of the verb or an affix/clitic that is phonologically bound to the verb.\\
The term "standard negation" refers to constructions that mark negation in declarative sentences involving dynamic (not-stative) verbal predicates.\\
Morphemes that attach (become phonologically bound) to other constituents, not verbs only, do not count.\\
Clitic boundaries are marked in the glosses by an equals sign: "=".\\
Affix boundaries are marked in the glosses by a dash: "-".\\
Separate words (i.e., particles that are not phonologically bound to other words) are separated from other words by spaces.\\
Choose one of the 2 following options: 1, 0.\\
Reply with 1 if standard negation in the language <...> can be marked by an affix, clitic or modification of the verb.\\
Reply with 0 if standard negation in <...> cannot be marked by an affix, clitic or modification of the verb.\\
1. Provide the reasoning for the chosen option.\\
2. After the reasoning, output the word "Conclusion:" and the chosen option at the end of your response.}
\end{quote}

\end{figure*}

\begin{figure*}
Prompt for Polar Questions (partially based on Grambank chapters related to strategies for marking polar questions):

\begin{quote}
\textit{Please determine all possible strategies for forming polar questions (yes-no questions) in the language <...>.\\
Consider neutral polar questions only (non-neutral, or leading, polar questions indicate that the speaker expects a particular response).\\
The 7 strategies for forming polar questions are the following: Interrogative intonation only, Interrogative word order, Clause-initial question particle, Clause-final question particle, Clause-medial question particle, Interrogative verb morphology, Tone.\\
Clitic boundaries are marked in the glosses by an equals sign: "=".\\
Affix boundaries are marked in the glosses by a dash: "-".\\
Separate words (e. g. particles that are not phonologically bound to other words) are separated from other words by spaces.\\
For this feature, count interrogative clitics as particles if they can be bound to other constituents in the sentence, not to the verb only.\\
Interrogative morphemes that can be phonologically bound to the verb only are counted as interrogative verbal morphology.\\
If a morpheme (for example, clitic or particle) can follow any constituent, which can be in various positions within the clause, including the clause-final position, code 1 for both "Clause-medial question particle" and "Clause-final question particle".\\
If a morpheme (for example, clitic or particle) can precede any constituent, which can be in various positions within the clause, including the clause-initial position, code 1 for both "Clause-initial question particle" and "Clause-medial question particle".\\
For each strategy, code 1 if it is present in the described language; code 0 if it is absent in the language.\\
Example of the output for a language that marks polar questions either with interrogative intonation only or with a clause-final interrogative particle:\\
"Interrogative intonation only: 1, Interrogative word order: 0, Clause-initial question particle: 0, Clause-final question particle: 1, Clause-medial question particle: 0, Interrogative verb morphology: 0, Tone: 0"\\
1. Provide the reasoning for the chosen option.\\
2. After the reasoning, output the word "Conclusion:" and the chosen option at the end of your response.}
\end{quote}

\end{figure*}

\begin{figure*}

Prompt for Number of Cases (partially based on WALS 49A):

\begin{quote}
\textit{Please determine the number of cases in the language <...>.\\
The term "cases" in the context of this feature refers to productive case paradigms of nouns.\\
Reply with one of the 9 following options: No morphological case-marking, 2 cases, 3 cases, 4 cases, 5 cases, 6-7 cases, 8-9 cases, 10 or more cases, Exclusively borderline case-marking.\\
The feature value "Exclusively borderline case-marking" refers to languages which have overt marking only for concrete (or "peripheral", or “semantic”) case relations, such as locatives or instrumentals.\\
Categories with pragmatic (non-syntactic) functions, such as vocatives or topic markers, are not counted as case even if they are morphologically integrated into case paradigms.\\
Genitives are counted as long as they do not encode categories of the possessum like number or gender as well, if they do not show explicit adjective-like properties. Genitives that may take additional case affixes agreeing with the head noun case ("double case") are not regarded as adjectival.\\
1. Provide the reasoning for the chosen option.\\
2. After the reasoning, output the word "Conclusion:" and the chosen option at the end of your response.}
\end{quote}

\includegraphics[width=1.63\linewidth]{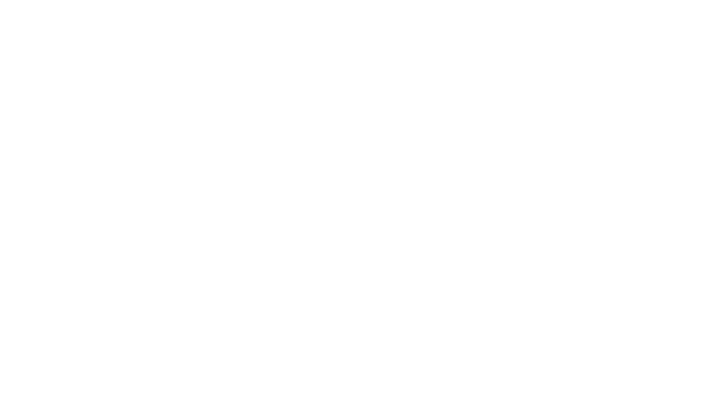}

\end{figure*}

\clearpage

\section{Benchmark for RAG: Prompt Details}
\label{sec:prompt_details}

\subsection{Wikipedia Summaries}

\begin{table}[ht]
\centering
\begin{tabular}{l|l}
\multicolumn{1}{c|}{\textbf{Feature}} & \multicolumn{1}{c}{\textbf{Wikipedia Title}} \\ \hline
\rowcolor[HTML]{FFFFFF} 
WALS 81A & Word order\tablefootnote{\url{https://en.wikipedia.org/w/index.php?title=Word_order&oldid=1240489972}} \\
\rowcolor[HTML]{FFFFFF} 
GB 107 & Affirmation and negation\tablefootnote{\url{https://en.wikipedia.org/w/index.php?title=Affirmation_and_negation&oldid=1202783032}} \\
\rowcolor[HTML]{FFFFFF} 
WALS 49A & Grammatical case\tablefootnote{\url{https://en.wikipedia.org/w/index.php?title=Grammatical_case&oldid=1238822129}} \\
\rowcolor[HTML]{FFFFFF} 
WALS 116A* & Yes-no question\tablefootnote{\url{https://en.wikipedia.org/w/index.php?title=Yes\%E2\%80\%93no_question&oldid=1236424936}} \\
\end{tabular}
\caption{\label{wikisummaries}
    The Wikipedia pages we chose to take their summaries as definitions for the typological features (versions as of September 3rd, 2024).
  }
\end{table}

The fourth paragraph from the "Grammatical case" summary was not included in the prompts due to containing examples of languages and their numbers of cases. If it was included in the prompt, we would have essentially provided the RAG system with an answer to a part of the given task.

\subsection{Base Prompts}
Due to the multilabel classification on the linguistic domain already posing a potential significant challenge for the RAG pipeline, we simplified the annotation guidelines for the features pertaining to interrogative particles compared to the ones presented in Grambank.

Firstly, in our version of the feature, the particles do not necessarily have to be dedicated to marking polar questions; the main purpose of the particle may be marking content questions, as long as using it to mark polar questions is possible as well. The motivation for this adjustment is the fact that the retriever and reranker in the RAG pipeline will retrieve information related to polar questions only based on the query, and it is irrational to expect information on polar questions to appear in the retrieved paragraphs.

Secondly, we deemed it unnecessary for a clause-medial particle to have the middle of the clause as its most common placement; a particle is considered clause-medial if it can appear in the clause-medial position in at least one example.

Another simplification is omission of two other features related to polar questions in Grambank: GB286, GB297. V-not-V constructions and double-marking with both a particle and verbal morphology occur in our benchmark in isolated examples only. 

Despite the simplifications, we impose the requirement on assigning clitics to the particle category if they can attach to anything in their vicinity, and to the verb morphology category if they can be phonologically bound to the verb only, in order for the task to test the capabilities of the RAG pipeline in regard to distinguishing morphology from syntax. GB291 (marking polar questions by means of tone) has been retained for a similar purpose: distinguishing morphology from prosody.

\subsection{Modifications for Chain-of-Thought}
We modified each chapter of the WALS and Grambank for the Chain-of-Thought prompts in the following way according to our specifications:
We eliminated all mentions and examples of languages that already exist in the
respective benchmarks. For instance, (1) has been modified to obtain (2) due to Oneida being present in the Number of Cases benchmark:

\begin{quote} (1) \textit{This excludes, for example, the "locative" suffixes in Oneida (Iroquoian; Ontario) from being counted as case, since they can derive body-part nouns which may occur in all semantically permitted syntactic positions (i.e. not only as locational adverbials). (WALS, Chapter 49A: Number of Cases)}
\end{quote}
\begin{quote} (2) \textit{For example, "locative" suffixes that can derive body-part nouns which may occur in all semantically permitted syntactic positions (i.e. not only as locational adverbials), are not counted as case.}
\end{quote}

Additionally, we removed mentions of the WALS map, since each prompt for the RAG pipeline is formulated as determining a feature value instead of placing it on a map. We eliminated duplicated information (i.e., left only one mention about disregarding leading polar questions across all chapters in GB257 – GB 291); changed the guidelines according to our simplified definitions of clause-initial/clause-medial/clause-final particles; and included the explanation of when clitics are regarded as particles and when they are considered verbal morphology.

\clearpage

\begin{table*}[H]
\section{Benchmark for RAG: Distribution of Values}
\label{sec:value_distrib}
        \vspace*{1.5 cm}
    \begin{minipage}{.5\linewidth}
        \centering
        \begin{tabular}{lc}
        \hline
        \multicolumn{2}{c}{\textbf{Word Order (WALS 81A)}} \\ \hline
        \multicolumn{1}{l|}{No dominant order} & 40 \\
        \multicolumn{1}{l|}{No mention} & 12 \\
        \multicolumn{1}{l|}{OSV} & 1 \\
        \multicolumn{1}{l|}{OVS} & 1 \\
        \multicolumn{1}{l|}{SOV} & 59 \\
        \multicolumn{1}{l|}{SVO} & 25 \\
        \multicolumn{1}{l|}{VOS} & 4 \\
        \multicolumn{1}{l|}{VSO} & 6 \\ \hline
        \multicolumn{1}{l|}{\textbf{Total}} & 148 \\ \hline
        \end{tabular}
        
        \vspace*{0.7 cm}
        
        \begin{tabular}{lc}
        \hline
        \multicolumn{2}{c}{\textbf{Number of Cases (WALS 49A)}} \\ \hline
        \multicolumn{1}{l|}{{No case marking}} & 55 \\
        \multicolumn{1}{l|}{{Excl. borderline case marking}} & 21 \\
        \multicolumn{1}{l|}{{2 cases}} & 11 \\
        \multicolumn{1}{l|}{{3 cases}} & 3 \\
        \multicolumn{1}{l|}{{4 cases}} & 5 \\
        \multicolumn{1}{l|}{{5 cases}} & 6 \\
        \multicolumn{1}{l|}{{6-7 cases}} & 22 \\
        \multicolumn{1}{l|}{{8-9 cases}} & 13 \\
        \multicolumn{1}{l|}{{10 or more cases}} & 12 \\ \hline
        \multicolumn{1}{l|}{\textbf{Total}} & 148 \\ \hline
        \end{tabular}
        
        \vspace*{0.5 cm}
        
        \begin{tabular}{lc}
        \hline
        \multicolumn{2}{c}{\textbf{Standard Negation (GB 107)}} \\ \hline
        \multicolumn{1}{l|}{0} & 65 \\
        \multicolumn{1}{l|}{1} & 83 \\ \hline
        \multicolumn{1}{l|}{\textbf{Total}} & 148 \\ \hline
        \end{tabular}
    \end{minipage}%
    \begin{minipage}{.5\linewidth}
      \centering
        \begin{tabular}{lc}
        \hline
        \multicolumn{2}{c}{\textbf{Polar Questions (WALS 116A*)}} \\ \hline
        \multicolumn{2}{c}{\textbf{Clause-initial particle}} \\ \hline
        \multicolumn{1}{l|}{0} & 20 \\
        \multicolumn{1}{l|}{1} & 128 \\ \hline
        \multicolumn{2}{c}{\textbf{Clause-medial particle}} \\ \hline
        \multicolumn{1}{l|}{0} & 132 \\
        \multicolumn{1}{l|}{1} & 16 \\ \hline
        \multicolumn{2}{c}{\textbf{Clause-final particle}} \\ \hline
        \multicolumn{1}{l|}{0} & 100 \\
        \multicolumn{1}{l|}{1} & 48 \\ \hline
        \multicolumn{2}{c}{\textbf{Verb morphology}} \\ \hline
        \multicolumn{1}{l|}{0} & 125 \\
        \multicolumn{1}{l|}{1} & 23 \\ \hline
        \multicolumn{2}{c}{\textbf{Interrogative word order}} \\ \hline
        \multicolumn{1}{l|}{0} & 144 \\
        \multicolumn{1}{l|}{1} & 4 \\ \hline
        \multicolumn{2}{c}{\textbf{Interrogative intonation only}} \\ \hline
        \multicolumn{1}{l|}{0} & 96 \\
        \multicolumn{1}{l|}{1} & 52 \\ \hline
        \multicolumn{2}{c}{\textbf{Tone}} \\ \hline
        \multicolumn{1}{l|}{0} & 141 \\
        \multicolumn{1}{l|}{1} & 7 \\ \hline
        \multicolumn{1}{l|}{\textbf{Total}} & 148 \\ \hline
        \end{tabular}
    \end{minipage}
    \newline
    \vspace*{0.2 cm}
    \newline
    \caption{\label{distribution of classes}
        Distribution of values in each feature (total = 148). The imbalance of classes should not be mitigated, because the benchmark, serving as a representative language sample, is intended to reflect the natural distribution of values found in languages around the world. If the grammar does not mention the feature, the following values are assigned to it: \textit{No mention} for \textbf{WALS 81A}; \textit{No case marking} for \textbf{WALS 49A}; \textit{0} for \textbf{GB 107}; \textit{0} for each of the sub-features for \textbf{WALS 116A*}. We make the assumption that a lack of mention in the descriptive material most likely indicates a lack of the phenomenon in the language (apart from the Word Order feature, since one of the seven logically possible categories must inevitably be present).
    }
    \vspace*{5 cm}
\end{table*}

\clearpage

\begin{figure*}
\section{Benchmark for Rerankers: Ablation}
\label{sec:ablation}

Additionally to the main experiment, we conducted an ablation study: how would the RAG pipeline perform if it had access to a near-optimal retriever instead of the combination of BM25 and rerankers, i.e., how much of the error is attributable to the generation component of Retrieval-Augmented Generation?

In order to conduct this ablation study, we annotated the pages in the grammar documents where the relevant information for the feature is contained (i.e., sufficient information for a linguist to unambiguously infer the value of the feature). Since this approach is only applicable to grammars that actually contain the relevant information, we ran the ablation on
136 grammars for WALS 81A, 146 grammars for GB 107, 121 grammars for WALS 116A, and 140 grammars for WALS 49A (out of 148), the same subsets as for the baseline runs. 

We used the manually retrieved pages as the retrieved component in RAG, resulting in two columns on the right in Table~\ref{abl}, and recalculated the metrics for the four RAG configurations on the same subsets as used for the ablation. The baseline metrics are identical to the ones in Table~\ref{rag_results} due to identical subsets of grammars. 

The results indicate that for features that have a high baseline performance (most likely due to having an extreme class imbalance), e.g, \textit{Interrogative Word Order} and \textit{Tone}, involving a human to retrieve the paragraphs may harm the classification quality. 

For WALS 49A (the feature hypothesized to be the most difficult one to classify) it is the Chain-of-Thought component that makes the difference between harming the classification performance and improving it; however, the F1 scores for this feature remain low, not exceeding 0.56.

\end{figure*}

\begin{table*}[H]
\centering
\small

\begin{tabular}{l|l|l|r|r|r|rlc|c}
 & \textbf{F1 average} & \multicolumn{1}{c|}{\textbf{Baseline}} & \multicolumn{1}{c|}{\textbf{Wiki}} & \multicolumn{1}{c|}{\textbf{\begin{tabular}[c]{@{}c@{}}Wiki\\ +CoT\end{tabular}}} & \multicolumn{1}{c|}{\textbf{\begin{tabular}[c]{@{}c@{}}Re-\\ ranker\end{tabular}}} & \multicolumn{1}{c}{\textbf{\begin{tabular}[c]{@{}c@{}}Rer.\\ +CoT\end{tabular}}} &  & \multicolumn{1}{l|}{\textbf{Human}} & \multicolumn{1}{l}{\textbf{\begin{tabular}[c]{@{}l@{}}Human\\ +CoT\end{tabular}}} \\ \cline{1-7} \cline{9-10} 
 & \textbf{micro} & \cellcolor[HTML]{AFC9CF}0.5551 ± 0.0359 & \cellcolor[HTML]{84ACB4}0.7353 & \cellcolor[HTML]{82AAB3}0.7426 & \cellcolor[HTML]{80A9B2}0.7500 & \cellcolor[HTML]{79A4AE}\textbf{0.7794} &  & \cellcolor[HTML]{CFE2F3}\textbf{0.7868} & \cellcolor[HTML]{CFE2F3}\textbf{0.7868} \\
 & \textbf{macro} & \cellcolor[HTML]{F1F6F7}0.2812 ± 0.0276 & \cellcolor[HTML]{8BB0B8}\textbf{0.7060} & \cellcolor[HTML]{A7C4CA}0.5866 & \cellcolor[HTML]{A7C4CA}0.5880 & \cellcolor[HTML]{A4C2C8}0.6000 &  & \cellcolor[HTML]{F4CCCC}\textbf{0.5198} & \cellcolor[HTML]{F4CCCC}0.4949 \\
\multirow{-3}{*}{\textbf{WALS 81A}} & \textbf{weighted} & \cellcolor[HTML]{B4CDD2}0.5328 ± 0.0337 & \cellcolor[HTML]{82ABB4}0.7398 & \cellcolor[HTML]{80A9B2}0.7499 & \cellcolor[HTML]{83ABB4}0.7391 & \cellcolor[HTML]{7BA5AF}\textbf{0.7721} &  & \cellcolor[HTML]{CFE2F3}0.7838 & \cellcolor[HTML]{CFE2F3}\textbf{0.7939} \\ \cline{1-7} \cline{9-10} 
 & \textbf{micro} & \cellcolor[HTML]{AAC6CC}0.5747 ± 0.0350 & \cellcolor[HTML]{95B7BF}0.6644 & \cellcolor[HTML]{95B7BF}0.6644 & \cellcolor[HTML]{90B4BC}0.6849 & \cellcolor[HTML]{87AEB7}\textbf{0.7192} &  & \cellcolor[HTML]{CFE2F3}0.8151 & \cellcolor[HTML]{CFE2F3}\textbf{0.8425} \\
 & \textbf{macro} & \cellcolor[HTML]{AAC6CC}0.5740 ± 0.0354 & \cellcolor[HTML]{ACC7CD}0.5679 & \cellcolor[HTML]{A9C5CB}0.5814 & \cellcolor[HTML]{A2C0C7}0.6100 & \cellcolor[HTML]{95B7BF}\textbf{0.6644} &  & \cellcolor[HTML]{CFE2F3}0.8038 & \cellcolor[HTML]{CFE2F3}\textbf{0.8304} \\
\multirow{-3}{*}{\textbf{GB 107}} & \textbf{weighted} & \cellcolor[HTML]{ABC6CC}0.5724 ± 0.0361 & \cellcolor[HTML]{A5C2C9}0.5958 & \cellcolor[HTML]{A2C0C7}0.6069 & \cellcolor[HTML]{9CBCC3}0.6334 & \cellcolor[HTML]{90B4BC}\textbf{0.6829} &  & \cellcolor[HTML]{CFE2F3}0.8102 & \cellcolor[HTML]{CFE2F3}\textbf{0.8366} \\ \cline{1-7} \cline{9-10} 
 & \textbf{micro} & \cellcolor[HTML]{D5E3E6}0.3986 ± 0.0238 & \cellcolor[HTML]{BCD2D7}0.5000 & \cellcolor[HTML]{B7CED4}0.5214 & \cellcolor[HTML]{B4CCD1}\textbf{0.5357} & \cellcolor[HTML]{BCD2D7}0.5000 &  & \cellcolor[HTML]{F4CCCC}0.5000 & \cellcolor[HTML]{CFE2F3}\textbf{0.5429} \\
 & \textbf{macro} & \cellcolor[HTML]{FFFFFF}0.2216 ± 0.0158 & \cellcolor[HTML]{C5D8DC}\textbf{0.4655} & \cellcolor[HTML]{C9DBDF}0.4461 & \cellcolor[HTML]{CDDDE1}0.4299 & \cellcolor[HTML]{D4E2E5}0.4031 &  & \cellcolor[HTML]{F4CCCC}0.4126 & \cellcolor[HTML]{CFE2F3}\textbf{0.4737} \\
\multirow{-3}{*}{\textbf{WALS 49A}} & \textbf{weighted} & \cellcolor[HTML]{E2EBED}0.3453 ± 0.0237 & \cellcolor[HTML]{BAD1D5}0.5088 & \cellcolor[HTML]{B4CCD2}0.5341 & \cellcolor[HTML]{B2CBD1}\textbf{0.5409} & \cellcolor[HTML]{BCD2D6}0.5023 &  & \cellcolor[HTML]{F4CCCC}0.5082 & \cellcolor[HTML]{CFE2F3}\textbf{0.5579} \\ \cline{1-7} \cline{9-10} 
 & \textbf{micro} & \cellcolor[HTML]{C8DADE}0.4504 ± 0.0396 & \cellcolor[HTML]{6C9BA6}0.8347 & \cellcolor[HTML]{5E929D}\textbf{0.8926} & \cellcolor[HTML]{6A9AA4}0.8430 & \cellcolor[HTML]{6496A1}0.8678 &  & \cellcolor[HTML]{CFE2F3}0.9174 & \cellcolor[HTML]{CFE2F3}\textbf{0.9256} \\
 & \textbf{macro} & \cellcolor[HTML]{CFDEE2}0.4240 ± 0.0456 & \cellcolor[HTML]{6C9BA6}0.8346 & \cellcolor[HTML]{5E929D}\textbf{0.8924} & \cellcolor[HTML]{6A9AA4}0.8429 & \cellcolor[HTML]{6496A1}0.8677 &  & \cellcolor[HTML]{CFE2F3}0.9169 & \cellcolor[HTML]{CFE2F3}\textbf{0.9251} \\
\multirow{-3}{*}{\textbf{\begin{tabular}[c]{@{}l@{}}interrog.\\ intonation\\ only\end{tabular}}} & \textbf{weighted} & \cellcolor[HTML]{D3E1E4}0.4068 ± 0.0484 & \cellcolor[HTML]{6B9BA6}0.8352 & \cellcolor[HTML]{5D929D}\textbf{0.8929} & \cellcolor[HTML]{699AA4}0.8433 & \cellcolor[HTML]{6396A1}0.8681 &  & \cellcolor[HTML]{CFE2F3}0.9178 & \cellcolor[HTML]{CFE2F3}\textbf{0.9260} \\ \cline{1-7} \cline{9-10} 
 & \textbf{micro} & \cellcolor[HTML]{4C8692}0.9653 ± 0.0076 & \cellcolor[HTML]{46828F}\textbf{0.9917} & \cellcolor[HTML]{488390}0.9835 & \cellcolor[HTML]{488390}0.9835 & \cellcolor[HTML]{4A8491}0.9752 &  & \cellcolor[HTML]{B4A7D6}\textbf{0.9917} & \cellcolor[HTML]{F4CCCC}0.9835 \\
 & \textbf{macro} & \cellcolor[HTML]{95B7BF}0.6641 ± 0.0460 & \cellcolor[HTML]{528A96}\textbf{0.9423} & \cellcolor[HTML]{5D919D}0.8957 & \cellcolor[HTML]{6395A0}0.8707 & \cellcolor[HTML]{6D9DA7}0.8269 &  & \cellcolor[HTML]{F4CCCC}\textbf{0.9264} & \cellcolor[HTML]{F4CCCC}0.8707 \\
\multirow{-3}{*}{\textbf{\begin{tabular}[c]{@{}l@{}}interrog.\\ word order\end{tabular}}} & \textbf{weighted} & \cellcolor[HTML]{4D8793}0.9611 ± 0.0064 & \cellcolor[HTML]{45818E}\textbf{0.9922} & \cellcolor[HTML]{478390}0.9851 & \cellcolor[HTML]{488390}0.9835 & \cellcolor[HTML]{498491}0.9765 &  & \cellcolor[HTML]{F4CCCC}\textbf{0.9912} & \cellcolor[HTML]{F4CCCC}0.9835 \\ \cline{1-7} \cline{9-10} 
 & \textbf{micro} & \cellcolor[HTML]{86ADB5}0.7264 ± 0.0402 & \cellcolor[HTML]{5C909C}0.9008 & \cellcolor[HTML]{5A8F9B}0.9091 & \cellcolor[HTML]{528A96}\textbf{0.9421} & \cellcolor[HTML]{568C98}0.9256 &  & \cellcolor[HTML]{CFE2F3}\textbf{0.9587} & \cellcolor[HTML]{F4CCCC}0.9339 \\
 & \textbf{macro} & \cellcolor[HTML]{B2CBD0}0.5446 ± 0.0502 & \cellcolor[HTML]{719FA9}0.8127 & \cellcolor[HTML]{6B9BA5}0.8385 & \cellcolor[HTML]{5C919C}\textbf{0.8972} & \cellcolor[HTML]{6496A1}0.8679 &  & \cellcolor[HTML]{CFE2F3}\textbf{0.9266} & \cellcolor[HTML]{F4CCCC}0.8889 \\
\multirow{-3}{*}{\textbf{\begin{tabular}[c]{@{}l@{}}clause-\\ initial\\ particle\end{tabular}}} & \textbf{weighted} & \cellcolor[HTML]{83ABB4}0.7371 ± 0.0337 & \cellcolor[HTML]{5C919C}0.8987 & \cellcolor[HTML]{598F9B}0.9100 & \cellcolor[HTML]{518A96}\textbf{0.9427} & \cellcolor[HTML]{558C98}0.9263 &  & \cellcolor[HTML]{CFE2F3}\textbf{0.9591} & \cellcolor[HTML]{F4CCCC}0.9362 \\ \cline{1-7} \cline{9-10} 
 & \textbf{micro} & \cellcolor[HTML]{BED3D8}0.4909 ± 0.0409 & \cellcolor[HTML]{87AEB7}0.7190 & \cellcolor[HTML]{77A3AD}0.7851 & \cellcolor[HTML]{8BB1B9}0.7025 & \cellcolor[HTML]{77A3AD}\textbf{0.7851} &  & \cellcolor[HTML]{CFE2F3}\textbf{0.8017} & \cellcolor[HTML]{CFE2F3}0.8017 \\
 & \textbf{macro} & \cellcolor[HTML]{C1D5D9}0.4808 ± 0.0415 & \cellcolor[HTML]{87AEB7}0.7190 & \cellcolor[HTML]{78A4AD}0.7826 & \cellcolor[HTML]{8BB1B9}0.7025 & \cellcolor[HTML]{78A4AD}\textbf{0.7826} &  & \cellcolor[HTML]{CFE2F3}\textbf{0.8000} & \cellcolor[HTML]{CFE2F3}0.7977 \\
\multirow{-3}{*}{\textbf{\begin{tabular}[c]{@{}l@{}}clause-\\ final\\ particle\end{tabular}}} & \textbf{weighted} & \cellcolor[HTML]{C4D8DC}0.4661 ± 0.0428 & \cellcolor[HTML]{87AEB6}0.7195 & \cellcolor[HTML]{77A3AD}0.7874 & \cellcolor[HTML]{8BB1B9}0.7030 & \cellcolor[HTML]{77A3AD}\textbf{0.7874} &  & \cellcolor[HTML]{CFE2F3}\textbf{0.8038} & \cellcolor[HTML]{CFE2F3}0.8035 \\ \cline{1-7} \cline{9-10} 
 & \textbf{micro} & \cellcolor[HTML]{92B5BD}0.6760 ± 0.0367 & \cellcolor[HTML]{6A9AA4}0.8430 & \cellcolor[HTML]{60939E}\textbf{0.8843} & \cellcolor[HTML]{6A9AA4}0.8430 & \cellcolor[HTML]{6697A2}0.8595 &  & \cellcolor[HTML]{CFE2F3}\textbf{0.9091} & \cellcolor[HTML]{CFE2F3}0.9008 \\
 & \textbf{macro} & \cellcolor[HTML]{BCD2D7}0.5011 ± 0.0544 & \cellcolor[HTML]{75A2AB}0.7961 & \cellcolor[HTML]{6B9BA5}\textbf{0.8374} & \cellcolor[HTML]{75A2AB}0.7961 & \cellcolor[HTML]{73A0AA}0.8049 &  & \cellcolor[HTML]{CFE2F3}\textbf{0.8671} & \cellcolor[HTML]{CFE2F3}0.8606 \\
\multirow{-3}{*}{\textbf{\begin{tabular}[c]{@{}l@{}}clause-\\ medial\\ particle\end{tabular}}} & \textbf{weighted} & \cellcolor[HTML]{95B7BF}0.6644 ± 0.0355 & \cellcolor[HTML]{6899A3}0.8502 & \cellcolor[HTML]{5F939E}\textbf{0.8857} & \cellcolor[HTML]{6899A3}0.8502 & \cellcolor[HTML]{6597A2}0.8621 &  & \cellcolor[HTML]{CFE2F3}\textbf{0.9085} & \cellcolor[HTML]{CFE2F3}0.9021 \\ \cline{1-7} \cline{9-10} 
 & \textbf{micro} & \cellcolor[HTML]{7CA6AF}0.7678 ± 0.0185 & \cellcolor[HTML]{7BA6AF}0.7686 & \cellcolor[HTML]{719FA9}\textbf{0.8099} & \cellcolor[HTML]{73A1AA}0.8017 & \cellcolor[HTML]{6F9EA8}\textbf{0.8182} &  & \cellcolor[HTML]{CFE2F3}0.8760 & \cellcolor[HTML]{CFE2F3}\textbf{0.8843} \\
 & \textbf{macro} & \cellcolor[HTML]{C8DADE}0.4533 ± 0.0311 & \cellcolor[HTML]{8BB1B9}0.7026 & \cellcolor[HTML]{83ABB4}0.7361 & \cellcolor[HTML]{85ADB5}0.7279 & \cellcolor[HTML]{80A9B2}\textbf{0.7506} &  & \cellcolor[HTML]{CFE2F3}0.8188 & \cellcolor[HTML]{CFE2F3}\textbf{0.8235} \\
\multirow{-3}{*}{\textbf{\begin{tabular}[c]{@{}l@{}}interrog.\\ verb\\ morphology\end{tabular}}} & \textbf{weighted} & \cellcolor[HTML]{8AB0B8}0.7102 ± 0.0160 & \cellcolor[HTML]{76A3AC}0.7894 & \cellcolor[HTML]{6E9DA7}0.8226 & \cellcolor[HTML]{709EA8}0.8157 & \cellcolor[HTML]{6C9CA6}\textbf{0.8311} &  & \cellcolor[HTML]{CFE2F3}0.8819 & \cellcolor[HTML]{CFE2F3}\textbf{0.8877} \\ \cline{1-7} \cline{9-10} 
 & \textbf{micro} & \cellcolor[HTML]{558C98}0.9273 ± 0.0128 & \cellcolor[HTML]{588E99}0.9174 & \cellcolor[HTML]{508895}0.9504 & \cellcolor[HTML]{4E8793}0.9587 & \cellcolor[HTML]{4C8692}\textbf{0.9669} &  & \cellcolor[HTML]{F4CCCC}0.9504 & \cellcolor[HTML]{B4A7D6}\textbf{0.9669} \\
 & \textbf{macro} & \cellcolor[HTML]{B9D0D5}0.5147 ± 0.0573 & \cellcolor[HTML]{80A9B2}0.7500 & \cellcolor[HTML]{6F9EA8}0.8199 & \cellcolor[HTML]{6698A2}\textbf{0.8572} & \cellcolor[HTML]{6899A4}0.8484 &  & \cellcolor[HTML]{F4CCCC}0.8199 & \cellcolor[HTML]{CFE2F3}\textbf{0.8662} \\
\multirow{-3}{*}{\textbf{tone}} & \textbf{weighted} & \cellcolor[HTML]{598F9A}0.9104 ± 0.0121 & \cellcolor[HTML]{548C97}0.9309 & \cellcolor[HTML]{4E8894}0.9555 & \cellcolor[HTML]{4C8693}0.9637 & \cellcolor[HTML]{4C8692}\textbf{0.9669} &  & \cellcolor[HTML]{F4CCCC}0.9555 & \cellcolor[HTML]{CFE2F3}\textbf{0.9689}
\end{tabular}
\caption{\label{abl}
    F1 scores for the ablation experiments (on the right) and the four initial RAG configurations + baseline (on the left) on the same subset of grammars as the ablation experiments. The best F1 value in each row in the main table and the best F1 value for the better of the two ablation experiments are shown in \textbf{bold}.  Each of the two ablation experiments was executed only once. Blue cells in the ablation experiments indicate \textbf{better} performance, red cells indicate \textbf{worse} performance, and purple cells indicate \textbf{identical} performance compared to the the best non-human retriever in the same row.
  }
\end{table*}

\clearpage

\begin{figure*}[ht]
Most features demonstrate an increase in F1 scores when compared to non-human retrievers.

The greatest improvement in quality was achieved on the balanced binary value, GB 107: +0.166 macro F1 compared to the best RAG configuration.

Overall, a conclusion can be made that while the task of machine reading on the linguistic domain is still not fully resolved, the task of information retrieval within the linguistic domain is similarly important.

    \vspace*{24 cm}
\end{figure*}

\end{document}